\documentclass{article}

\usepackage[final]{corl_2026} 

\usepackage{amsmath}
\usepackage{amssymb}
\usepackage{mathtools}
\usepackage{amsthm}

\usepackage{caption}
\usepackage{hyperref}
\usepackage{url}
\usepackage{xcolor}
\usepackage{graphicx}   
\usepackage{booktabs}
\usepackage{multirow}
\usepackage[table]{xcolor} 
\usepackage{wrapfig}
\usepackage{enumitem}
\usepackage{array}

\usepackage[T1]{fontenc}
\usepackage{textcomp}

\usepackage{tikz}
\usetikzlibrary{tikzmark}
\usepackage{algorithm}

\usepackage{inconsolata} 
\usepackage{listings}
\lstdefinestyle{pythonic}{
    language=Python,
    basicstyle=\ttfamily\footnotesize,
    keywordstyle=\color{blue}\bfseries,         
    commentstyle=\color{green!50!black},       
    stringstyle=\color{orange},                
    emph={torch,nn,F,self}, emphstyle=\color{purple}\bfseries,
    showstringspaces=false,
    breaklines=true
}

\definecolor{softpeach}{HTML}{FFEDE5} 
\definecolor{coral}{HTML}{ff6937} 
\definecolor{linkblue}{HTML}{0000EE}

\newcommand{\scissoremoji}{\raisebox{-0.1ex}{\includegraphics[height=0.8em]{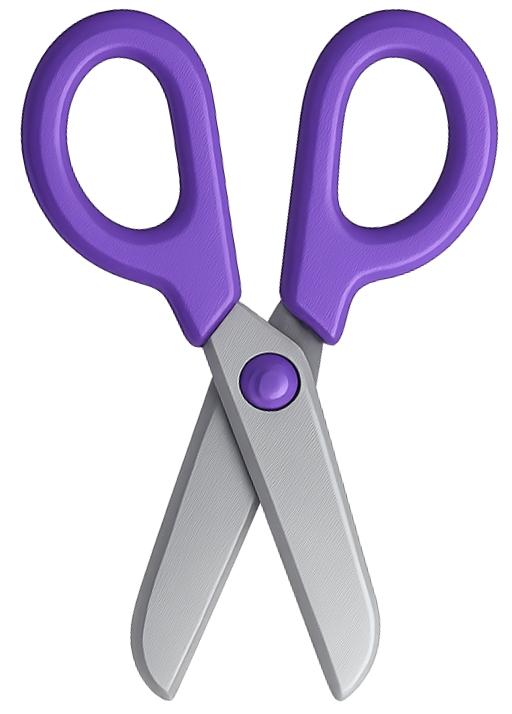}}}

\newcommand{\method}[0]{GLUESTICK}

\definecolor{FAIorange}{HTML}{F35937}
\newcommand{\emphFAI}[1]{\textbf{\textcolor{black}{#1}}}
\usepackage{hyperref}
\usepackage{subcaption}

\usepackage{xcolor}
\usepackage{pifont}

\newcommand{\cmark}{\textcolor{green!60!black}{\ding{51}}}
\newcommand{\xmark}{\textcolor{red!75!black}{\ding{55}}}

\usepackage[capitalize,noabbrev]{cleveref}

\theoremstyle{plain}

\theoremstyle{definition}

\theoremstyle{remark}

\title{\textbf{Don't Run with Scissors \scissoremoji:} \textnormal{Pruning Breaks VLA Models but They Can Be Recovered}}

\author{
  \normalfont\mdseries
  \textbf{Jason Jabbour}$^{1,2,\star}$ \quad
  \textbf{Dong-Ki Kim}$^{1}$ \quad
  \textbf{Max Smith}$^{1}$ \quad
  \textbf{Jay Patrikar}$^{1}$ \quad
  \textbf{Radhika Ghosal}$^{2}$ \\[1.3em]
  \textbf{Youhui Wang}$^{2}$ \quad
  \textbf{Ali Agha}$^{1}$ \quad
  \textbf{Vijay Janapa Reddi}$^{2}$ \quad
  \textbf{Shayegan Omidshafiei}$^{1}$ \\[1.6em]
  \normalfont\mdseries
  $^{1}$FieldAI \qquad $^{2}$Harvard University
  \vspace{-1em}
}

\begin{document}
\maketitle

{\renewcommand{\thefootnote}{$\star$}%
  \footnotetext{Work conducted during Jason Jabbour’s internship at FieldAI. Contact: \texttt{jasonjabbour@g.harvard.edu}}}

\begin{abstract} 
Vision-Language-Action (VLA) models are a promising paradigm for field robotics, but their deployment is bottlenecked by strict edge SWaP-C constraints. Compression techniques such as quantization, distillation, and pruning are well-established for LLMs and VLMs. Pruning, however, remains particularly underexplored for VLA models. We surprisingly find that pruning VLA models yields drastic performance degradation and increased safety violations. Through spectral analysis, we identify structural properties of VLA weights that help explain this fragility. Based on this insight, we introduce \textbf{\method}, a post-pruning recovery method that restores much of the original model's capability while preserving sparsity benefits. \method\, computes a one-time corrective term by interpolating between the dense and pruned models in weight space. At inference, each pruned layer applies this correction to recover lost performance with minimal overhead. Our approach requires no additional training and is agnostic to the pruning algorithm. Across diverse VLA architectures and domains, \method\, recovers up to 100\% of the success rate and restores safety to near-dense levels, while saving up to 5 GB of memory and achieving up to 1.75$\times$ speedup. Additional material can be found at: \href{https://gluestick-vla.github.io/}{\textcolor{linkblue}{\textbf{https://gluestick-vla.github.io/}}}.

\end{abstract}


\begin{figure*}[h!]
    \centering
    \includegraphics[width=1\textwidth]{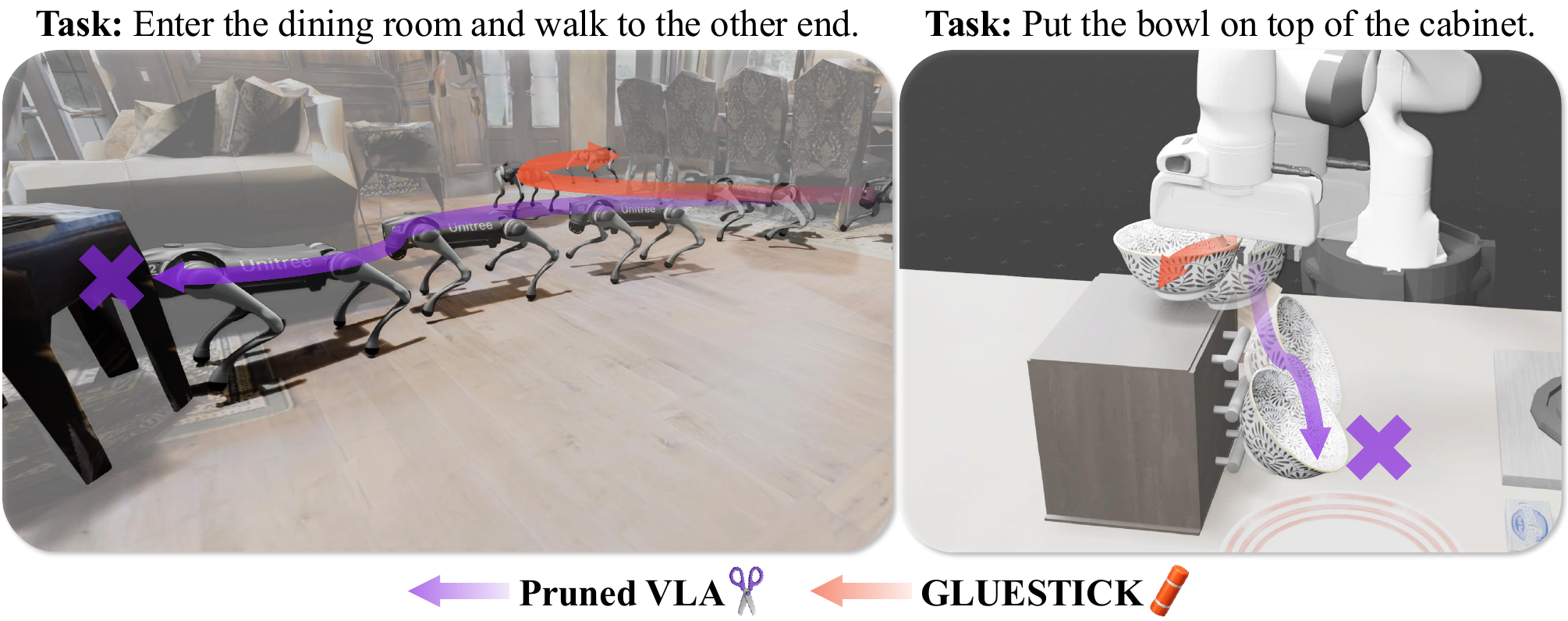}
    \caption{
    \textbf{VLA models break under pruning, and GLUESTICK fixes them.} Pruning methods unexpectedly cause task and safety failures in VLAs: colliding with an object in a navigation task \textbf{(left)}, or dropping a bowl in a manipulation task \textbf{(right)}. Our training-free post-pruning method, \textbf{\method{}}, restores the lost functionality by adding a lightweight correction computed purely in weight space, while largely retaining the memory and speed benefits of pruning.}
    \label{fig:teaser}
\end{figure*}

\vspace{-1em}
\section{Introduction}
\vspace{-.5em}
Field robots operate in environments such as subterranean mines and tunnels \cite{dixit2024step, miller2020mine, tranzatto2022cerberus, hudson2022heterogeneous, scherer2022resilient, ginting2024semantic}, disaster zones \cite{nagatani2013emergency, murphy2014disaster, delmerico2019current}, offshore platforms \cite{gehring2021anymal, khatib2016ocean}, and on the surface of other planets \cite{balaram2018mars, morrell2024robotic, patel2023towards}, where connectivity to off-board compute is unreliable or too slow for real-time control \cite{best2022resilient, agha2021nebula, chinchali2021network, alsolami2025evaluating}. As a result, field robots perform all computation onboard, on edge hardware governed by strict Size, Weight, Power, and Cost (SWaP-C) constraints \cite{neuman2022tiny, nguyen2019mavnet, ennasr2023low, krishnan2022roofline}. Recently, a new paradigm has emerged in robotics: Vision-Language-Action (VLA) models~\citep{Kim2024OpenVLA, li2024cogact, black2024pi_0, bjorck2025gr00t}.
Many current VLA models sit in the 7--8B parameter range, with some ranging up to 55--562B parameters \cite{zitkovich2023rt, driess2023palm}, and like the LLMs and VLMs that preceded them, VLAs follow scaling trends with the trend pointing upward~\citep{kaplan2020scaling, pearce2024scaling, agarwal2025cosmos}. The result is a footprint that exceeds the budget of widely deployed edge accelerators; for instance, OpenVLA 7B fails to load on the Jetson Orin NX entirely. For LLMs and VLMs, a mature toolbox of compression techniques has been developed: quantization \cite{dettmers2022gpt3}, distillation \citep{sanh2019distilbert}, and pruning \cite{sun2024simple}. In robotics, quantization \cite{Kim2024OpenVLA, zhang2026quantvla} and distillation \cite{ravichandran2025distilling, prasad2024consistency} have been applied to VLAs, but pruning has not, leaving its impact on task success and safety unknown.

While pruning preserves accuracy for LLMs and VLMs \citep{frantar2023sparsegpt, sun2024simple, koike2025mu}, we surprisingly find that applying the same methods to VLAs leads to catastrophic degradation. In our experiments, recent pruning algorithms reduced task success rate on manipulation tasks from 85.2\% to 0.0\% and on a navigation task from 43.0\% to 0.0\%, while also increasing the number of safety violations. Despite this failure, pruning is too valuable to abandon. Pruning is particularly valuable for robotics because it reduces both memory and inference latency by removing weights entirely \cite{mishra2021accelerating}. Quantization, distillation, and pruning are complementary and commonly composed together \cite{xu2024beyond, han2015deep, muralidharan2024compact}; removing any one from the toolbox places a ceiling on how compact the deployed model can be. Restoring pruning to the VLA compression toolbox is a central goal of this work.

To this end, we introduce \textbf{\method}, a post-pruning recovery method that restores signal lost during pruning while preserving the efficiency benefits of sparsity. Our spectral analysis reveals that VLA weights lack the concentrated spectral structure that makes LLMs and VLMs resilient to pruning, motivating \method's design. \method\, operates entirely in weight space, using the information discarded by pruning to nudge the model back toward more performant regions without any retraining. This is achieved by adding a lightweight correction term, computed from singular values in the gap between the dense and pruned weights. Our approach is pruning-agnostic and can be applied on top of any existing pruning algorithm. In doing so, \method\, restores up to 100\% of performance in some tasks, while maintaining the efficiency gains of structured sparsity by providing up to 5 GB memory savings and 1.75$\times$ speedup (Figure~\ref{fig:teaser}). 
We demonstrate that our method consistently recovers performance and improves safety across three VLA architectures, two widely used robotics benchmarks, multiple robot embodiments, and on real robotic systems. 

\textbf{Our contributions are as follows}:

\begin{itemize}[leftmargin=*, wide, labelindent=0pt, topsep=0pt, itemsep=.1pt,
  parsep=0pt]
    \item \textbf{Empirical evidence of pruning collapse in VLAs.} 
    We present the first systematic study showing that pruning, which is effective for LLMs and VLMs, causes near-complete collapse of the success rate in embodied VLA models, and an increase in safety violations.

    \item \textbf{Study of why VLAs differ from LLMs and VLMs under pruning.} 
    Through spectral analysis, we identify structural properties of VLAs that could make them fragile to pruning.

    \item \textbf{An effective, training-free recovery method.} 
    We propose \method, a post-pruning recovery algorithm that restores lost signal after pruning using a low-rank, lightweight correction in weight space. 
    \method\, is pruning-algorithm-agnostic and requires no retraining.

\end{itemize}

\vspace{-.5em}
\section{Related Work}
\vspace{-.5em}
\label{related_work}

\textbf{VLA Models.} Recent work has developed VLA models that unify perception, language understanding, and decision-making into a single policy mapping multimodal inputs directly to robot actions. 
Prominent systems include OpenVLA~\citep{Kim2024OpenVLA}, NaVILA~\citep{Cheng2025NaVILA}, RT~\citep{brohan2022rt,zitkovich2023rt}, the $\pi$ series~\citep{black2024pi_0,intelligence2025pi_}, PaLM-E~\citep{driess2023palm}, GR00T N1~\citep{bjorck2025gr00t}, and CogACT~\citep{li2024cogact}, all of which are large end-to-end transformer-based policies with billions of parameters.
The trend is toward richer inputs and outputs: SpatialVLA~\citep{qu2025spatialvla} incorporates 3D scene information, while MolmoAct~\citep{lee2025molmoact} and WorldVLA~\citep{Cen2025WorldVLA} extend outputs to include depth predictions or full world models, further growing model computational requirements. As these models scale, understanding how compression affects success and safety becomes critical.

\textbf{Pruning for LLMs and VLMs.} Pruning compresses models by setting a fraction of weights to zero~\citep{zhu2024survey}. Magnitude~\citep{han2015learning} removes small-magnitude weights, while Wanda~\citep{sun2024simple} scores connections by activation statistics on calibration inputs. Both are training-free and computationally efficient. When applied to LLMs, pruning achieves strong accuracy retention. On LLaMA-2-70B~\citep{touvron2023llama}, mean accuracy on the EleutherAI LM Harness~\citep{gao2021framework} decreases by only 7.13\% with Magnitude pruning and 2.94\% with Wanda~\citep{sun2024simple}, despite removing 50\% of weights under 2:4 structured sparsity. For VLMs, the impact is slightly more pronounced: \citet{koike2025mu} reported a 12.5\% accuracy drop on LLaVA-7B, and \citet{liang2025efficientllava} observed 9\%--30\% drops in LLaVA variants. No prior work has examined how pruning affects VLA success or safety, or how such impacts might be recovered.

\textbf{Weight Modifications and Safety.} Beyond pruning, other forms of weight adjustment have been widely studied, such as machine unlearning~\citep{golatkar2020eternal,liu2024towards} and fine-tuning \citep{hu2022lora}.  
These studies demonstrate that modifying weights can strongly affect downstream behavior: fine-tuning, for instance, has been shown to degrade LLM safety~\citep{yang2025asft,huangantidote}, motivating post-finetuning frameworks to preserve alignment~\citep{djuhera2025safemerge,hsu2024safe}.  
Taken together, this line of work suggests that while pruning has been validated on LLMs, weight adjustments cannot be assumed to be benign in all contexts.  
For VLAs, where degraded behavior has physical consequences, the safety implications of pruning remain unstudied. Recovering from such damage in weight space remains an open problem that we address.

\vspace{-.5em}
\section{Motivation: Necessity of Pruning for Edge Devices on Field Robots}
\vspace{-.5em}

The gap between VLA resource requirements and the constraints of edge hardware is severe. We detail the severity of that gap and the distinct role pruning plays in addressing it.

\textbf{Edge Hardware Constraints.} Since field robots perform computation onboard, the hardware itself must fit within tight SWaP-C budgets \cite{neuman2021robomorphic, jabbour2024generative, jabbour2022closing}.
Datacenter accelerators sit far outside this budget. An NVIDIA B200, for instance, consumes on the order of a kilowatt per GPU \cite{nvidia2025dgxB200}. A typical field robot's battery pack stores only a few hundred watt-hours, meaning a single datacenter GPU would drain it in minutes, before powering any motors or sensors \cite{jabbour2024generative}. Instead, the field uses a family of edge accelerators, which consume on the order of tens of watts and can run for hours on the same battery. The NVIDIA Jetson Orin NX \cite{nvidia_jetson_orin_nx_datasheet}, AGX Orin \cite{nvidia_agx_orin_technical_brief_2021}, and Jetson Thor \cite{nvidia_jetson_thor_2025} are the dominant platforms for onboard compute in field robotics~\citep{park2024quantization, liu2024omninxt, rey2025performance, nvidia_jetson_thor_blog_2025, robotreport_jetson_thor_2025}. The tradeoff is memory and compute. The Orin NX, for example, provides only 8 to 16 GB shared between CPU and GPU; the OS and every other process on the robot draw from the same pool that the robotic algorithms must fit into. This is far below server-grade hardware, which may pair eight HGX B200 GPUs, each with 180 GB of dedicated memory, with terabytes of host RAM. The compute gap is equally stark. The Orin NX delivers just 4.8 FP16 TFLOPs 
\cite{nvidia_jetson_orin_nx_datasheet}, while a single B200 provides 4,500 FP16 TFLOPs 
\cite{nvidia_b200_datasheet, cornell_b200_sm}, nearly 1,000$\times$ more compute. For field robots, which are real-time systems that must react quickly to dynamic environments, this compute deficit directly limits how fast control loops can run.

\textbf{Why Pruning Matters for Edge Robotics.}
Deploying VLAs on these devices requires reducing both memory footprint and inference latency. Pruning is especially attractive because it addresses both simultaneously by removing weights entirely rather than merely reducing their precision. When applied in hardware-friendly patterns such as structured ``N:M'' sparsity (e.g., 2:4), pruning reduces both parameter count and FLOPs. Modern GPUs exploit these patterns with specialized kernels that reduce memory traffic and multiply-accumulate operations (MACs); for example, NVIDIA's Sparse Tensor Cores and cuSPARSELt accelerate 2:4 sparse general matrix multiplications~\citep{mishra2021accelerating}. The reduced computation due to pruning not only enables acceleration and memory savings but also significantly cuts power consumption~\citep{han2016eie}. For devices operating under strict constraints on compute, memory, and energy, these benefits make pruning a critical tool for deploying VLAs at the edge. As discussed, quantization, distillation, and pruning are complementary; removing any one from the toolbox places a ceiling on how compact the deployed model can be. Our central goal is to bring pruning back into this toolbox for VLAs.

\begin{figure}[t]
\centering

\begin{minipage}[t]{0.45\linewidth}
\centering
\fontsize{7}{8}\selectfont

\renewcommand{\arraystretch}{1.2}
\setlength{\tabcolsep}{3pt}   

\begin{tabular}{p{0.22\linewidth} p{0.17\linewidth} r r c c}
\toprule
\textbf{Model} & \textbf{Method} & \textbf{Success} & \textbf{Unsafe} & \textbf{Fits} & \textbf{Accel.}\\
\midrule

 & Dense      & 85.2\% & 33.4\%  & \xmark & \xmark \\
\cmidrule(l){2-6}
  OpenVLA 7B       & Magnitude & 0.0\%  & 46.4\%  & \cmark & \cmark \\
\cmidrule(l){2-6}

         & Wanda     & 0.0\%  & 51.6\%  & \cmark & \cmark \\

\midrule
\midrule

  & Dense      & 43.0\% & 23.0\%  & \xmark  & \xmark \\
\cmidrule(l){2-6}

  NaVILA 8B     & Magnitude & 0.0\%  & 100.0\% & \cmark & \cmark \\
\cmidrule(l){2-6}

         & Wanda     & 0.0\%  & 46.0\%  & \cmark & \cmark \\
\bottomrule
\end{tabular}

\vspace{-0.1em}
\captionof{table}{
\textbf{Pruning impact on VLAs.}
Success ${=}$\% successful episodes;
Unsafe${=}$\% with a safety violation;
Fits${=}$ fits on Jetson Orin NX 16 GB;
Accel.${=}$ accelerated inference.}
\label{tab:pruning_collapse}
\end{minipage}
\hfill
\begin{minipage}[t]{0.53\linewidth}
\centering
\vspace{-5.8em}

\includegraphics[width=0.85\linewidth]{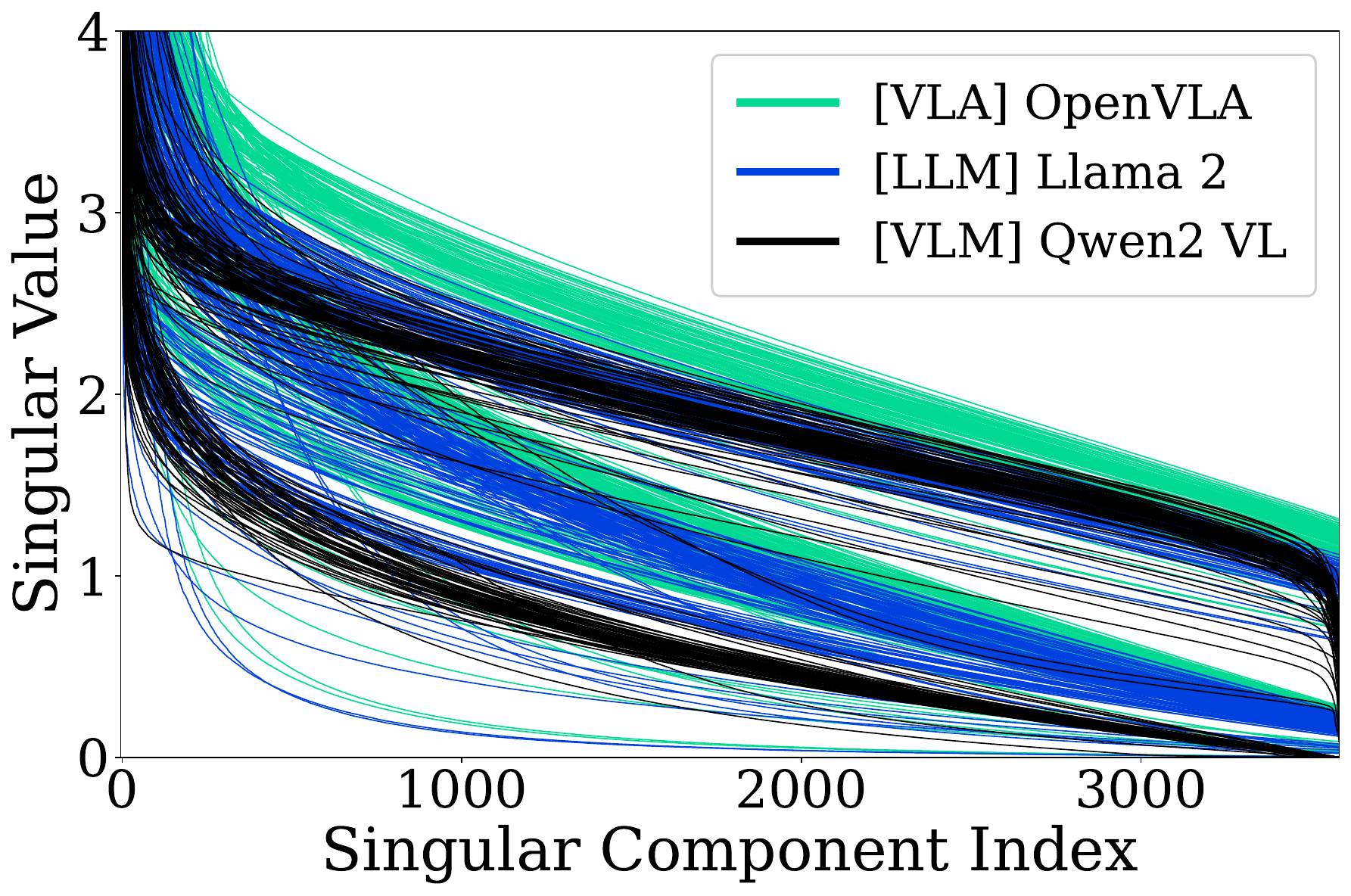}

\captionof{figure}{
\textbf{Singular value spectra of weights}.
VLA vs.\ LLM/VLM showing that VLA spectra are flatter and signal is more dispersed across the weight space.}
\label{fig:singular_value_spectra}
\end{minipage}

\vspace{-1em}

\end{figure}

\section{\method: Why Pruning Breaks VLA Models and How to Fix It}
In this section, we first present our surprising finding that pruning can cause catastrophic degradation in VLA success and safety.
We then introduce our new method, \textbf{GLUESTICK} (sin\textbf{G}ular va\textbf{LUE} \textbf{STI}t\textbf{C}hing), which glues pruned VLAs back to high task success rates and safe behaviors as in the original dense models. Additional details and pseudocode are available in Appendix \ref{app:gluestick_details}.

\subsection{Pruning Severely Degrades VLA Performance}
Pruning has substantially reduced memory usage in LLMs and VLMs, with minimal loss in accuracy (see Section~\ref{related_work}). 
At the same sparsity levels commonly used in LLMs and VLMs, we observe surprisingly different outcomes on popular VLAs.
Representative VLAs such as OpenVLA~\citep{Kim2024OpenVLA} and NaVILA~\citep{Cheng2025NaVILA}, when pruned with Magnitude or Wanda, drop in task success rate to 0\% (from 85.2\% and 43.0\%, respectively).
Pruning also increases unsafe behavior: OpenVLA's unsafe-episode rate rises from 33.4\% to 51.6\%, and NaVILA's from 23.0\% to 100.0\% (see Table~\ref{tab:pruning_collapse}).
These findings show that text and image-validated pruning techniques do not directly transfer to embodied control. Yet despite this, pruning remains essential as it allows models to fit on edge hardware and leverage sparse kernel acceleration (Table~\ref{tab:pruning_collapse}).

\subsection{Why Pruning Breaks VLAs}

\begin{wrapfigure}{r}{0.565\textwidth}
\vspace{-5.5em}

    \centering
    \includegraphics[width=0.56\textwidth]{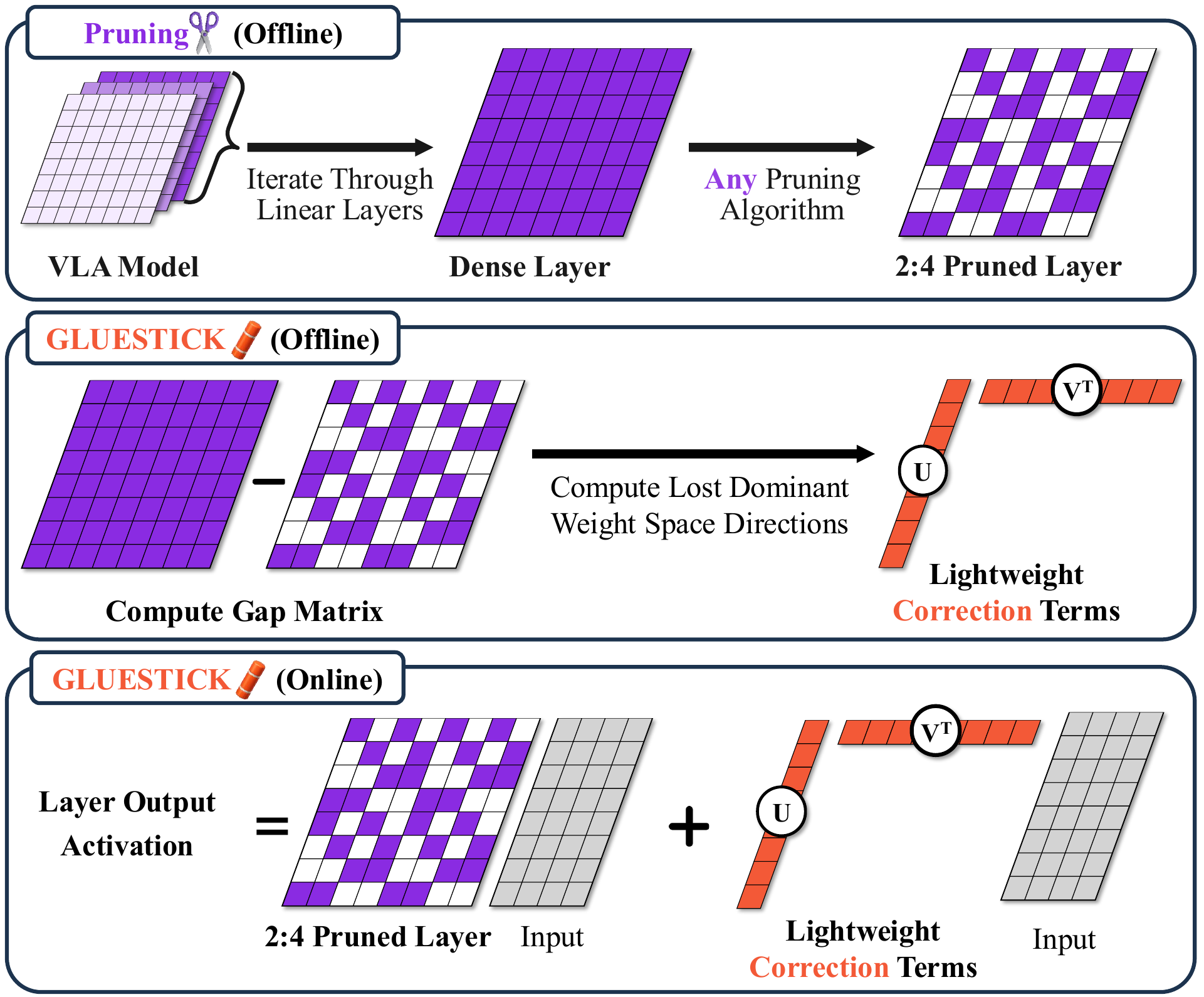}
    \caption{\textbf{Overview of \method.} \textbf{(Top)} A VLA model is pruned with a standard algorithm (e.g., Wanda) to enforce 2:4 sparsity in linear layers. \textbf{(Middle)} Offline, the gap between the dense and pruned weights is computed and dominant lost directions are extracted via SVD, yielding lightweight corrections. \textbf{(Bottom)} At inference, these correction terms are applied alongside the pruned weights, adding back lost signal.}
    \label{fig:method}
    \vspace{-1.5em}
\end{wrapfigure}

To explain why pruning degrades VLAs far more than LLMs or VLMs, we ask whether their weight-space properties differ.
Thus, we examine the singular value spectra of equal-sized layers within the language backbones of OpenVLA (VLA), LLaMA-2-7B (LLM) \citep{touvron2023llama}, and Qwen2-VL-7B-Instruct (VLM) \citep{bai2025qwen2, qwen2vl7binstruct}. We display these spectra in Figure~\ref{fig:singular_value_spectra}, where each line corresponds to the singular values of a single layer, obtained by computing the SVD of that layer’s weight matrix and plotting its full set of singular values. In LLM and VLM models, the spectra are more anisotropic, evidenced by a steep initial drop followed by a long tail, which concentrates energy in a few dominant directions.
This profile helps pruning, since removing small coefficients mainly trims low-energy directions while leaving the principal subspaces intact.
In contrast, VLA layers show a noticeably flatter decay, indicating energy spread across many directions.
In this regime, even small-magnitude coefficients contribute to important subspaces, so pruning discards useful signal distributed throughout the matrix (see Appendix~\ref{app:spectra} for additional details). 
Based on this insight, our method in the following section explores the recovery of this lost information within the weight space.

\subsection{The GLUESTICK Recovery Method for Pruned VLA Models}

The space of pruning configurations is combinatorial, making optimal selection of weights to remove intractable.
Heuristic methods such as Magnitude \cite{han2015learning} and Wanda \cite{sun2024simple} sidestep the global optimization by scoring individual weights and pruning by score.
While simple and efficient, these heuristics discard correlated weights under grouped sparsity constraints (e.g., 50\% sparsity with 2:4 or 4:8 groups), which could be especially harmful for VLA models.

We propose \textbf{\method}, a post-hoc, training-free recovery method that operates entirely in weight space and is agnostic to the pruning algorithm (see Figure~\ref{fig:method}).
\method\, requires only the original dense model and its pruned counterpart, and incurs a one-time offline cost; no additional training is required. 

Specifically, for each linear layer with dense weight matrix $W_\mathrm{dense}\in\mathbb{R}^{d_\mathrm{out}\times d_\mathrm{in}}$ and its pruned version $W_{\text{pruned}}$ (fixed, preserving the original 2:4/4:8 pattern), we define the \emph{gap matrix}:
\begin{equation}
    W_\mathrm{gap} = W_\mathrm{dense} - W_\mathrm{pruned},
\end{equation}
which captures lost information due to pruning. 
We then compute a truncated singular value decomposition (SVD) of the gap matrix:
\begin{equation}
    W_\mathrm{gap} = U \Sigma V^\top \approx U_r\Sigma_r V_r^\top,
\end{equation}
keeping the top $r$ singular components. 
By \citet{eckart1936approximation}, this is the \emph{best rank-$r$} approximation to $W_\mathrm{gap}$ in Frobenius norm.
For memory and speed, we fold the singular values into one term so that only two compact matrices need to be stored:
\begin{equation}
    A = U_r \Sigma_r \in \mathbb{R}^{d_\mathrm{out}\times r}, \quad 
    B = V_r \in \mathbb{R}^{d_\mathrm{in} \times r}.
\end{equation}

During inference, \method\, adds a lightweight correction around each pruned layer:
\begin{equation}
    h(x) = W_\mathrm{pruned} x + A(B^\top x),
\end{equation}
which re-injects the dominant lost directions at low cost while leaving $W_\mathrm{pruned}$ unchanged, thereby preserving the efficiency gains of structured sparsity with a minimal overhead addition from the correction term.
The extra compute from this correction is:
\begin{equation}
    \underbrace{W_\mathrm{pruned} x}_{\text{efficient sparse matmul}}
    +
    \underbrace{B^{\top} x}_{\mathcal{O}(d_{\mathrm{in}}\,r)}
    +
    \underbrace{A(\,\cdot\,)}_{\mathcal{O}(d_{\mathrm{out}}\,r)},
\end{equation}
or $\mathcal{O}((d_\mathrm{in}+d_\mathrm{out})r)$ on top of the sparse matrix-matrix multiplication (matmul), versus $\mathcal{O}(d_\mathrm{in}d_\mathrm{out})$ for the dense layer.
Our correction adds only $(d_\mathrm{in}+d_\mathrm{out})r$ extra parameters per layer, which is small compared to $d_\mathrm{in}d_\mathrm{out}$ in the dense case.
With $r \ll \min\{d_\mathrm{in}, d_\mathrm{out}\}$, \method\, preserves the efficiency gains of structured 50\% sparsity. 

We refer to our method as \method-$r$ to indicate the chosen value of $r$. 
We note that the parameter $r$ provides a dial between memory usage and recovery. 
Smaller values of $r$ favor memory savings, while larger values prioritize recovery. 
In practice, integrating \method{} into a model requires only a wrapper around pruned layers (see Appendix~\ref{app:gluestick_details} for details).

\begin{table*}[t]
\centering
\fontsize{8}{9}\selectfont
\setlength{\tabcolsep}{3pt}
\begin{minipage}[t]{0.49\textwidth}
\centering
\setlength{\tabcolsep}{4.5pt}
\begin{tabular}{@{}l rrrrr@{}}
\toprule
\textbf{Method} & \multicolumn{4}{c}{\textbf{LIBERO} ($\uparrow$)} & \textbf{Mean} ($\uparrow$)\\
\cmidrule(lr){2-5}
& \textbf{Spat.} & \textbf{Obj.} & \textbf{Goal} & \textbf{Long} \\
\midrule
Full Sparse       & $0.0$  & $13.4$ & $0.8$  & $0.0$  & $3.6$ \\
Sparse L.B.       & $17.3$ & $25.8$ & $20.5$ & $6.2$  & $17.5$ \\
\% Sparse L.B.    & $15.2$ & $25.2$ & $21.2$ & $5.7$  & $16.8$ \\
\emphFAI{\method} & \emphFAI{$\mathbf{54.0}$} & \emphFAI{$\mathbf{48.2}$} & \emphFAI{$\mathbf{46.1}$} & \emphFAI{$\mathbf{13.3}$} & \emphFAI{$\mathbf{40.4}$} \\
\bottomrule
\end{tabular}
\captionof{table}{%
    \textbf{Task success rate (\%).}
    Results averaged across OpenVLA and WorldVLA.
    \% Sparse L.B.\ uses the same VRAM as \method.
}
\label{tab:success_overview}
\end{minipage}%
\hfill
\begin{minipage}[t]{0.49\textwidth}
\centering
\setlength{\tabcolsep}{3pt}
\begin{tabular}{@{}l rrrrr@{}}
\toprule
\textbf{Method} & \multicolumn{4}{c}{\textbf{LIBERO} ($\downarrow$)} & \textbf{Mean} ($\downarrow$) \\
\cmidrule(lr){2-5}
& \textbf{Spat.} & \textbf{Obj.} & \textbf{Goal} & \textbf{Long} & \\
\midrule
Full Sparse       & $+18.2$ & $+23.0$ & $+1.4$ & $+11.6$ & $+13.6$ \\
Sparse L.B.       & $+9.0$  & $+13.2$ & $+1.8$ & $+9.4$  & $+8.4$ \\
\% Sparse L.B.    & $+5.6$  & $+14.4$ & $+4.6$ & $+5.2$  & $+7.5$ \\
\emphFAI{\method} & \emphFAI{$\mathbf{+0.2}$} & \emphFAI{$\mathbf{+1.2}$} & \emphFAI{$\mathbf{+0.0}$} & \emphFAI{$\mathbf{+4.6}$} & \emphFAI{$\mathbf{+1.5}$} \\
\bottomrule
\end{tabular}
\captionof{table}{%
    \textbf{Change in unsafe-episode rate (\%) relative to Full Dense.}
    Lower values indicate fewer episodes with safety violations.
}
\label{tab:safety_overview}
\end{minipage}
\end{table*}

\begin{table*}[t]
\vspace{-.6em}
\centering
\fontsize{8}{9}\selectfont
\setlength{\tabcolsep}{4pt}
\begin{minipage}[t]{0.6\textwidth}
\centering
\setlength{\tabcolsep}{10.5pt}
\begin{tabular}{@{}l rrrr@{}}
\toprule
\textbf{Method} & \textbf{Success} ($\uparrow$) & \textbf{PL (m)} ($\downarrow$) & \textbf{DG (m)} ($\downarrow$)\\
\midrule
Full Dense              & $43.0$ & $11.7$ & $5.9$ \\
Full Sparse             & $0.0$  & $17.6$ & $9.5$ \\
Sparse L.B.             & $23.0$ & $14.8$ & $8.5$ \\
\% Sparse L.B.        & $25.0$ & $14.9$ & $8.4$ \\
\emphFAI{\method}   & \emphFAI{$\mathbf{44.0}$} & \emphFAI{$\mathbf{11.9}$} & \emphFAI{$\mathbf{5.9}$} \\
\bottomrule
\end{tabular}
\captionof{table}{%
    \textbf{Navigation performance.}
    Success = task success rate (\%). \% Sparse L.B.\ uses the same VRAM as \method. PL $=$ Path Length. DG $=$ Final Distance to Goal. 
}
\label{tab:nav_performance}
\end{minipage}%
\hfill
\begin{minipage}[t]{0.38\textwidth}
\centering
\setlength{\tabcolsep}{30pt}
\begin{tabular}{@{}l r@{}}
\toprule
\textbf{Method} & $\boldsymbol{\Delta}$\textbf{Unsafe} ($\downarrow$) \\
\midrule
Full Dense              & $+0.0$ \\
Full Sparse             & $+23.0$ \\
Sparse L.B.             & $+2.0$ \\
\% Sparse L.B.        & $+12.0$ \\
\emphFAI{\method}   & \emphFAI{$\mathbf{-4.0}$} \\
\bottomrule
\end{tabular}
\captionof{table}{%
    \textbf{Navigation safety.}
    $\Delta$Unsafe is change in unsafe-episode rate (\%) relative to Full Dense; lower is better.
}
\label{tab:nav_safety}
\end{minipage}


\end{table*}

\section{Evaluation}

Our experimental setup spans two benchmarks covering distinct robotics domains: manipulation and navigation. 
We evaluate three different VLA models on these tasks, with results measured using both task success and safety metrics. We structure our results around key questions, first presenting main findings on \method\, then providing analysis through ablations and broader considerations.

\subsection{Experimental Setup}

\textbf{Environments.}
We evaluate on two benchmarks: LIBERO \citep{liu2023libero} for manipulation (see Figure~\ref{fig:teaser}, right), which has four task suites testing manipulation skills with a Franka Panda arm, and VLN-CE-Isaac \citep{Cheng2025NaVILA} for navigation (see Figure~\ref{fig:teaser}, left), which simulates legged robots traversing indoor environments. See Appendix~\ref{app:setting:environments} for additional details.

\textbf{Models.} We study three VLA models: OpenVLA \citep{Kim2024OpenVLA}, a 7B-parameter generalist VLA for manipulation; WorldVLA \citep{Cen2025WorldVLA}, a 7B manipulation-oriented VLA targeting long-horizon consistency; and NaVILA \citep{Cheng2025NaVILA}, an 8B-parameter navigation-focused VLA. See Appendix~\ref{app:setting:models} for details.

\textbf{Evaluation Metrics.} We evaluate on two axes: task success, defined as binary per-episode completion, and safety, defined as the absence of harm to the robot or its surroundings \citep{dulac2019challenges, geng2023a, morton2025safe}. For manipulation, we monitor risks such as joint-limit violations, collisions, unsafe object motion, and containment breaches; for navigation, we track collisions. See Appendix~\ref{app:methods:safety} for details.

\textbf{Baselines.} We compare against three pruning strategies: Full Sparse, Sparse Language Backbone, and a Memory-Matched Sparse Language Backbone (a subset of backbone layers pruned to match GLUESTICK's memory overhead). We use Wanda with 50\% 2:4 structured sparsity. Unless otherwise stated, \method~refers to \method-500. See Appendix~\ref{app:setting:baselines} for details.

\subsection{Main Results}

\textbf{Q1:} \emph{Does \method\, recover the task success rate for pruned VLAs?}

\method\, recovers the lost success rate by up to 100\%. On navigation, \method~recovers 100\% of the lost performance, matching the dense baseline at 44.0\% (Table~\ref{tab:nav_performance}). On manipulation, pruning collapses the average success rate across the four LIBERO suites to just 3.6\%. \method~recovers over 10$\times$ the pruned baseline's success rate, reaching 40.4\%, with recovery strongest in the Spatial suite (from 0.0\% to 54.0\%). \method~also outperforms the memory-matched pruning baseline by over 2$\times$ (Table~\ref{tab:success_overview}). See Appendix~\ref{app:results:success-rate} for more details.

\begin{table*}[t]
\centering
\fontsize{6}{7}\selectfont
\begin{tabular}{
  p{0.12\textwidth}
  p{0.16\textwidth}
  p{0.24\textwidth}
  >{\raggedleft\arraybackslash}p{0.19\textwidth}
  >{\raggedleft\arraybackslash}p{0.14\textwidth}
}
\toprule
\textbf{Model} & \textbf{Device} & \textbf{Method} & \textbf{Latency} ($\downarrow$) & \textbf{Speedup} ($\uparrow$) \\
\midrule
\multirow{5}{*}{OpenVLA}  & \multirow{5}{*}{NVIDIA Thor}
  & Full Dense              & $796.50 \pm 25.85$ ms  & ---            \\
& & Full Sparse             & $434.47 \pm 3.63$ ms   & $1.83\times$   \\
& & Sparse Language Backbone & $439.45 \pm 13.37$ ms & $1.81\times$   \\
& & GLUESTICK-200           & $453.58 \pm 31.29$ ms  & $1.75\times$   \\
& & GLUESTICK-500           & $537.49 \pm 28.19$ ms  & $1.48\times$   \\
\midrule
\multirow{4}{*}{WorldVLA} & \multirow{4}{*}{NVIDIA Thor}
  & Full Dense              & $2816.57 \pm 65.73$ ms & ---            \\
& & Sparse Language Backbone & $1488.44 \pm 57.27$ ms & $1.89\times$  \\
& & GLUESTICK-200           & $1691.05 \pm 69.35$ ms & $1.67\times$   \\
& & GLUESTICK-500           & $1860.56 \pm 48.23$ ms & $1.51\times$   \\
\midrule
\multirow{4}{*}{OpenVLA}  & \multirow{4}{*}{NVIDIA Orin NX}
  & Full Dense              & Out of Memory          & ---            \\
& & Sparse Language Backbone & $1546.37 \pm 57.32$ ms & Deploys$^\dagger$ \\
& & GLUESTICK-200           & $1790.99 \pm 42.15$ ms & Deploys$^\dagger$ \\
& & GLUESTICK-500           & $2085.96 \pm 58.50$ ms & Deploys$^\dagger$ \\
\bottomrule
\end{tabular}
\caption{
\textbf{Edge inference latency.} End-to-end inference on NVIDIA Jetson Thor and Orin NX. Pruning accelerates inference; \method\ retains speedup with minimal overhead. $\dagger$ Since dense exceeds device memory, speedup not compared but still accelerated; sparsity enables deployment.
}
\label{hardware_experiments_latency}
\vspace{-1em}
\end{table*}

\begin{table*}[t]
\centering
\fontsize{8}{9}\selectfont
\begin{minipage}[c]{0.48\textwidth}
\centering
\includegraphics[width=\textwidth]{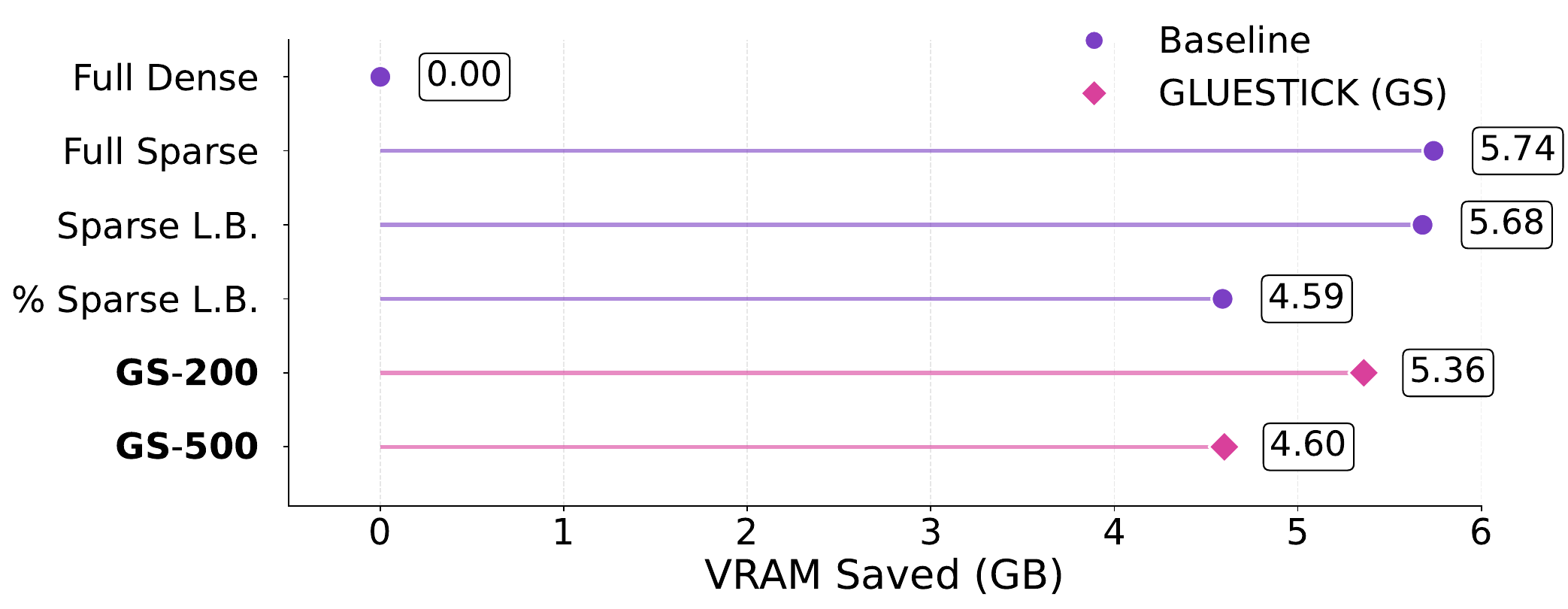}
\end{minipage}%
\hfill
\begin{minipage}[c]{0.48\textwidth}
\centering
\includegraphics[width=\textwidth]{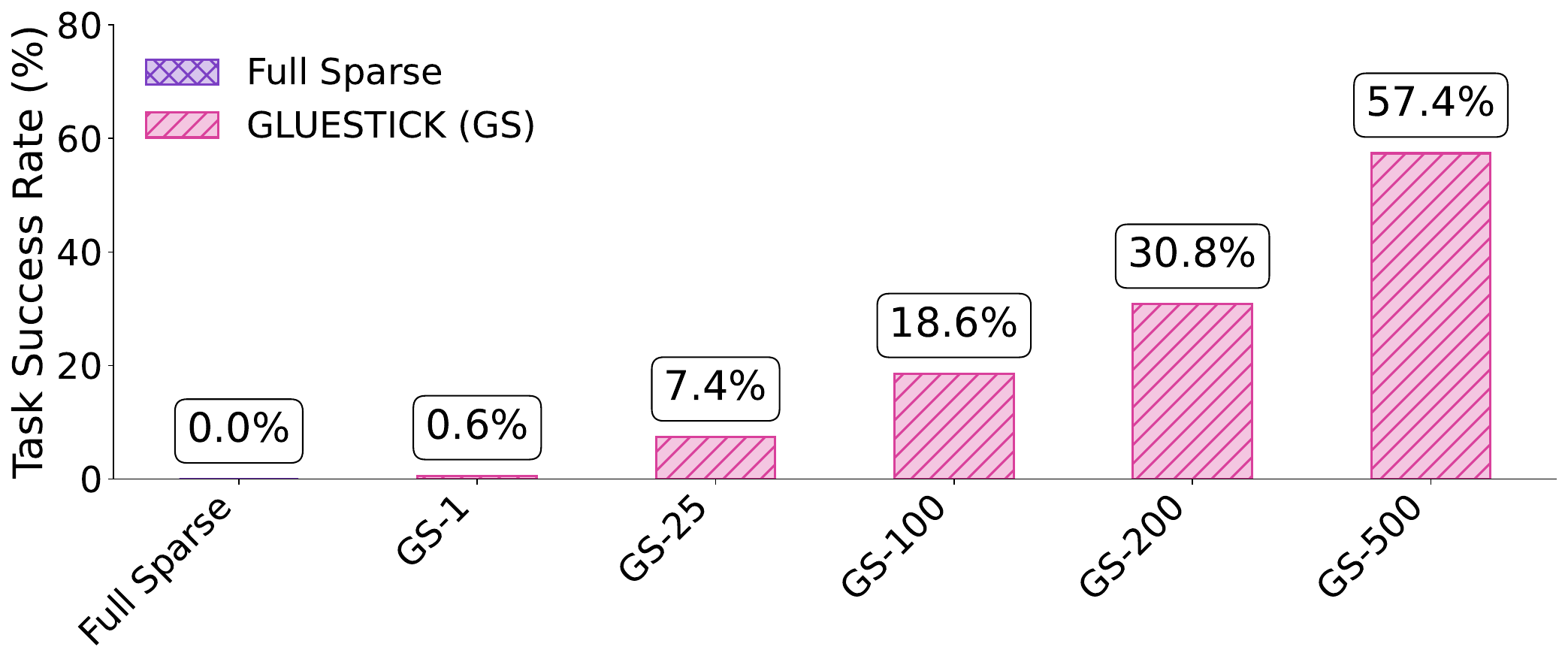}
\end{minipage}
\captionof{figure}{%
    \textbf{Memory Savings and Rank Tradeoff.} \textbf{Left:} VRAM savings relative to Full Dense. All methods use NaVILA. \textbf{Right:} Rank ablation. All methods use OpenVLA and LIBERO Spatial.
}
\label{fig:ram_and_rank}
\end{table*}

\method's recovery is particularly striking, restoring not only success but also trajectory efficiency. As seen in Table~\ref{tab:nav_performance}, \method~restores path length and final distance-to-goal to values nearly identical to those of the dense baseline. To appreciate the extent of this recovery, consider what pruning does without it. Pruning's failure is not a matter of less efficient paths; the robot's behavior is fundamentally altered. Without \method, pruned agents frequently veer into entirely different rooms rather than approaching the goal. After Full Sparse pruning, mean path length increases by over 50\% (11.7 to 17.6 m) and mean distance to goal by over 60\% (5.9 to 9.5 m).   This highlights that \method~not only mitigates pruning degradation but completely closes the gap to dense performance across both success and path-quality metrics. See Appendix~\ref{app:results:navigation} for details.

\textbf{Q2:} \emph{How well does \method\ restore the safety of pruned VLAs?}

Across both manipulation and navigation, \method~preserves the safety profile of the dense policy, remaining near parity despite aggressive pruning.
Tables~\ref{tab:safety_overview} and~\ref{tab:nav_safety} report changes in unsafe-episode rate relative to the dense baseline in manipulation and navigation, respectively.
\method~yields $89\%$ and $100\%$ fewer unsafe episodes compared to the Full Sparse model for manipulation and navigation, respectively. That is, \method~maintains the safety profile of the original dense models. These results indicate that \method\ restores dominant weight-space directions that carry both task-relevant and safety-critical signal, thereby preserving the safety of VLA models.

\textbf{Q3:} \emph{Does \method~preserve the inference speedup of pruning?}

We evaluate inference latency on two edge devices: the NVIDIA Jetson Thor and the Jetson Orin NX (Table~\ref{hardware_experiments_latency}). On Thor, pruning with sparse kernels accelerates end-to-end inference by up to 1.83$\times$--1.89$\times$. \method\, adds only minimal overhead and still achieves a 1.75$\times$
 speedup for OpenVLA and 1.67$\times$
for WorldVLA. Even \method-500, which provides stronger task recovery, maintains a 1.48$\times$ to 1.51$\times$ speedup. On the Orin NX, the dense OpenVLA model cannot run at all, exceeding the device's memory limit. Conversely, \method{}-200 successfully deploys on the Orin NX with only 245\,ms of overhead relative to the sparse baseline. This overhead is minimal compared to the alternative: without \method, the choice on the Orin NX is between a model that cannot load and a sparse model with severely degraded performance. See Appendix~\ref{inference_speedup}.

\begin{table*}[t]
\centering
\fontsize{8}{9}\selectfont
\setlength{\tabcolsep}{11pt}

\begin{tabular}{l rrrrr}
\toprule
\textbf{Method} &
\textbf{Dense} &
\textbf{Dense/Int8} &
\textbf{Sparse} &
\textbf{Sparse/Int8} &
\textbf{Sparse/Int8/\method} \\
\midrule
Success Rate (\%) & 85.2 & 84.8 & 31.2 & 31.8 & 62.2 \\
\bottomrule
\end{tabular}

\caption{\textbf{\method{} composes with quantization.} Evaluating OpenVLA on LIBERO Spatial.}
\label{tab:prune_quant_gluestick_int8}
\end{table*}

\begin{table*}[t]
\centering

\begin{minipage}[t]{0.48\linewidth}
\centering
\fontsize{8}{9}\selectfont

\begin{tabular}{
  p{0.39\linewidth}
  >{\raggedleft\arraybackslash}p{0.08\linewidth}
  >{\raggedleft\arraybackslash}p{0.08\linewidth}
  >{\raggedleft\arraybackslash}p{0.08\linewidth}
  >{\raggedleft\arraybackslash}p{0.08\linewidth}
}

\toprule
\textbf{Method} &
\textbf{0\%} &
\textbf{30\%} &
\textbf{50\%} &
\textbf{70\%} \\
\midrule
Wanda & 85.2 & 80.8 & 31.8 & 1.2 \\
Wanda~+~\method & -- & \textbf{85.8} & \textbf{60.2} & \textbf{24.6} \\
\bottomrule
\end{tabular}
\caption{
    \textbf{\method-500's recovery of VLA success rates across sparsities.} Evaluated on LIBERO Spatial. Results for the pruned language backbone of OpenVLA using Wanda.
}
\label{tab:sparsity_ablation}
\end{minipage}
\hfill
\begin{minipage}[t]{0.495\linewidth}
\centering
\fontsize{8}{9}\selectfont
\begin{tabular}{
  p{0.39\linewidth}
  >{\raggedleft\arraybackslash}p{0.08\linewidth}
  >{\raggedleft\arraybackslash}p{0.08\linewidth}
  >{\raggedleft\arraybackslash}p{0.08\linewidth}
  >{\raggedleft\arraybackslash}p{0.08\linewidth}
}

\toprule
\textbf{Method} &
\textbf{0\%} &
\textbf{30\%} &
\textbf{50\%} & 
\textbf{70\%} \\
\midrule
Wanda & 93.88 & 93.38 & 88.16 & 21.30\\
Wanda~+~\method & -- & -- & \textbf{92.95} & \textbf{90.89} \\
\bottomrule
\end{tabular}
\caption{
    \textbf{\method-500's ability to recover pruned VLM performance.}
    Evaluated on 1000 DocVQA examples using Wanda-pruned Qwen2-VL-7B-Instruct VLM. Values show ANLS.
}
\label{tab:qwen2vl_docvqa}
\end{minipage}
\vspace{-1.5em}
\end{table*}

\textbf{Q4:} \emph{What are \method's memory savings and how does $r$ affect recovery–memory trade-off?}

\method~delivers substantial memory savings. As shown in Figure~\ref{fig:ram_and_rank} (left), \method-200 saves 5.36 GB of VRAM, offering memory savings within 0.38 GB of the fully sparse model, while recovering nearly the full dense success rate (Table~\ref{tab:navila_memory}). By comparison, a fully sparse NaVILA

\begin{wraptable}{r}{0.5\linewidth}
\centering
\fontsize{8}{9}\selectfont
\begin{tabular}{
  p{0.50\linewidth}
  >{\raggedleft\arraybackslash}p{0.14\linewidth}
  >{\raggedleft\arraybackslash}p{0.14\linewidth}
}
\toprule
\textbf{Method} &
\textbf{2:4} &
\textbf{4:8} \\
\midrule
Magnitude & 11.0 & 25.4 \\
Magnitude~+~\method{} & \textbf{48.8} & \textbf{61.8} \\
\bottomrule
\end{tabular}
\caption{\textbf{\method{} is pruning-algorithm and sparsity-pattern agnostic.} Magnitude-pruned OpenVLA language backbone evaluated on LIBERO Spatial (values show success \%).}
\label{tab:pruning_agnostic}
\vspace{-1.5em}
\end{wraptable}

model achieves the maximum memory savings of 5.74 GB but suffers a 0\% success rate. This illustrates the trade-off that \method\, exposes through a single hyperparameter $r$ as shown in Figure~\ref{fig:ram_and_rank} (right), increasing $r$ improves success-rate recovery at the cost of additional memory, giving practitioners a direct dial between efficiency and task performance. See Appendix~\ref{app:results:success-rate} and \ref{app:results:navigation} for more details.

\vspace{-0.5em}
\subsection{Analysis}
\vspace{-0.5em}

\textbf{Q5:} \emph{How compatible is \method~with other compression methods?}

\method~composes well with other compression techniques. To demonstrate this, we combine \texttt{Int8} quantization with pruning and \method. The results in Table \ref{tab:prune_quant_gluestick_int8} show that quantization has minimal impact on the success rate of both dense and sparse models; however, pruning degradation remains severe even after quantization. \method~recovers performance to 62.2\% even on top of quantization, achieving our goal of making pruning a viable tool in the VLA compression toolbox.

We also show that \method{} can be combined with VLA-specific compression techniques such as token pruning. Token pruning reduces the number of input tokens, while \method{} operates on pruned weights, making the two complementary rather than competing. We ran experiments combining \method{} with VLA-Pruner~\cite{liu2025vlapruner} on NVIDIA Thor and found that their speedups compound: \method{}'s 1.48$\times$ speedup rises to 1.64$\times$, with an additional 0.12\,GB of VRAM savings (VLA-Pruner targets latency rather than memory reduction). VLA-Pruner costs a 7\% decrease in success due to token removal. This further shows \method{}’s composability.

\begin{figure*}[t]
    \centering
    \includegraphics[width=1\textwidth]{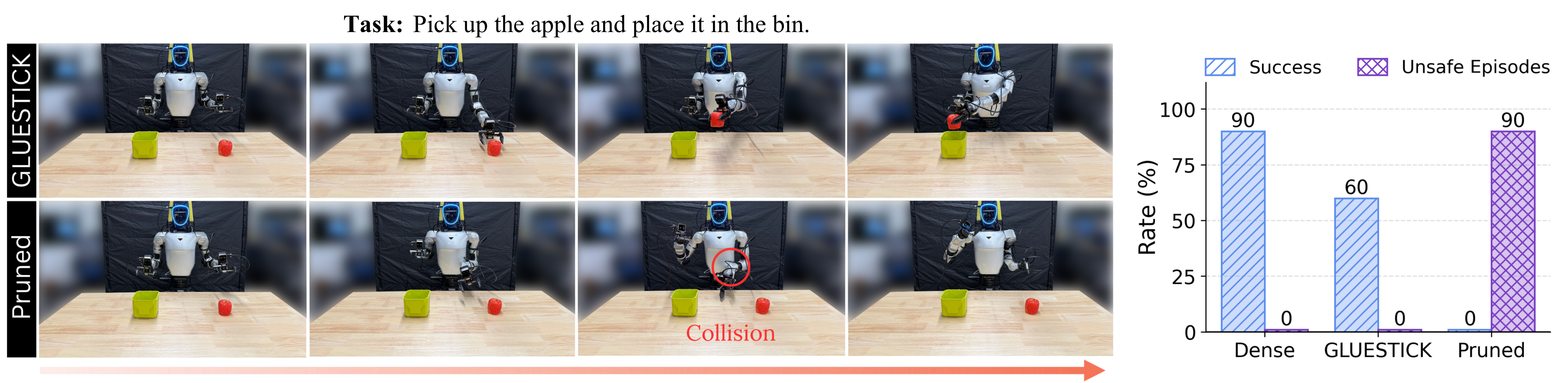}
    \caption{\textbf{Real-World Manipulation on Unitree G1.} With the pruned model, the G1's arm collides with its own body and the task fails, while with \method{} the G1 successfully places the apple in the bin (\textbf{left}). Pruning results in 90\% unsafe episodes and 0\% task success, while \method{} fully restores safety and reaches 60\% success (\textbf{right}).}
    \vspace{-1em}
    \label{fig:g1}
\end{figure*}

\begin{figure*}[th!]
    \centering
        \fontsize{8}{9}\selectfont
    \includegraphics[width=\linewidth]{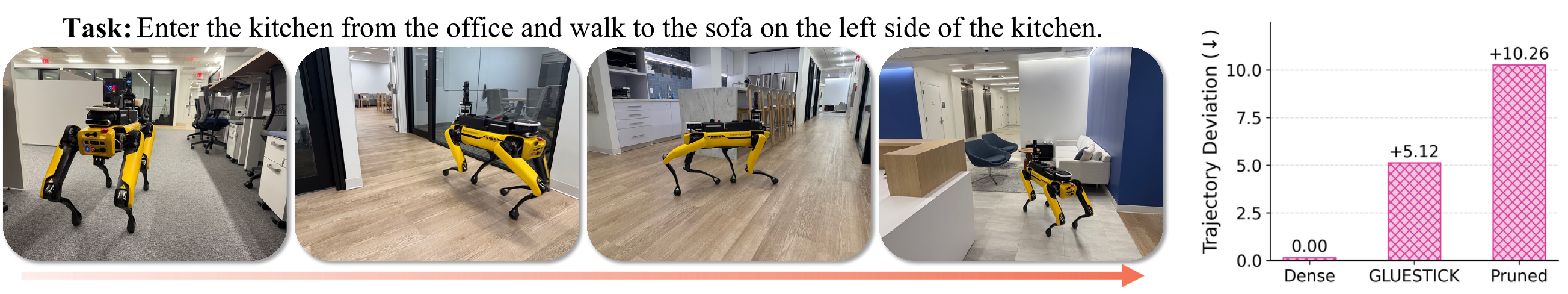}

    \vspace{0cm}
    \caption{\textbf{Real-World Navigation on Boston Dynamics Spot.} Our real Spot robot navigating in a real office environment \textbf{(left)}. \method~reduces pruned NaVILA deviation by 50\% \textbf{(right)}.}
    \vspace{-2em}
\label{fig:realworld}

\end{figure*}

\textbf{Q6:} \emph{Is \method{} agnostic to the pruning algorithm and sparsity pattern?}

To test whether \method{} generalizes beyond the Wanda 2:4 setting used in our main results, we apply it to a Magnitude-pruned language backbone of OpenVLA under both 2:4 and 4:8 sparsity and evaluate on LIBERO Spatial (see Table~\ref{tab:pruning_agnostic}). \method{} improves success from 11.0\% to 48.8\% under 2:4 sparsity and from 25.4\% to 61.8\% under 4:8 sparsity, indicating that it is effective across both pruning algorithms and sparsity patterns.

\textbf{Q7:} \emph{How does \method~ perform at varying VLA sparsity levels?}

We evaluate \method's capability to recover pruned VLA models under different unstructured sparsity ratios in Table~\ref{tab:sparsity_ablation}. In this experiment, we prune the language backbone in OpenVLA and observe that GLUESTICK recovers performance effectively across sparsity levels.

\textbf{Q8:} \emph{Does \method~generalize to restore pruned VLMs?}

We evaluate \method's capability to recover the performance of a pruned VLM under various sparsity ratios in Table~\ref{tab:qwen2vl_docvqa}. We find strong generality, as \method~recovers VLM performance even at a harsh 70\% sparsity, significantly improving ANLS from 21.30 to 90.89.

\textbf{Q9:} \emph{How does \method{} perform on real-world robots?}

We evaluate \method{} on two real-world robotic platforms, a Unitree G1 humanoid for manipulation and a Boston Dynamics Spot for navigation. For manipulation, our setup (Figure~\ref{fig:g1}, left) deploys GR00T~N1.7~\cite{nvidia2026gr00tn17} on the G1, fine-tuned to pick up an apple and place it in a bin. We prune both the vision and language backbones of the VLA using Magnitude pruning at 50\% 2:4 sparsity. We run 10 trials per model and apply \method{}-1000 to all pruned layers, reporting results in Figure~\ref{fig:g1} (right). Pruning causes the model to fail completely, reducing success from 90\% to 0\% and producing unsafe behavior in 90\% of episodes, with safety as defined in Appendix~\ref{app:methods:safety}. Interestingly, in nearly every episode, the pruned model exhibited self-collisions, collisions with the table, containment breaches in which the end effector moved beyond the table's edges, or a combination of these. We also observed the arms twisting into contorted, unnatural poses. In contrast, \method{} fully restores safety and reaches 60\% success.

For navigation, we record observations from a real Spot robot maneuvering through a real office environment (Figure~\ref{fig:realworld}, left). We prune NaVILA using Wanda at 50\% 2:4 sparsity and replay the recorded observations through each model, measuring trajectory deviation, defined as the cumulative L1 difference in linear and angular velocity commands relative to the Full Dense model. \method{} reduces this deviation by 50\% relative to the pruned model (Figure~\ref{fig:realworld}, right).

\vspace{-.5em}
\section{Conclusion}
\vspace{-.5em}

We presented the first systematic study of pruning VLA models and showed that pruning severely degrades both task success and safety. To address this, we introduced \method{}, a training-free, pruning-agnostic, and practical post-pruning recovery method that reintroduces lost directions due to pruning, restoring performance and safety while retaining efficiency. Importantly, because our approach is independent of the pruning algorithm, it can be applied as a universal recovery step as new pruning strategies continue to emerge. We hope this work lays the foundation for compression techniques that make VLAs deployable on resource-constrained robots.

\textbf{Limitations.}
We designed GLUESTICK with a single hyperparameter so that applications can tune it based on their needs, but this introduces an inherent trade-off between success, memory, and latency. We also applied a uniform rank ($r$) across all layers. Future work includes reducing this trade-off and exploring adaptive rank selection across layers for further efficiency and recovery.

\acknowledgments{We thank Rumi Loghmani for help with the humanoid demonstration.  Jason Jabbour is supported by the NSF GRFP under Grant No.~DGE-2140743. Any opinions, findings, and conclusions or recommendations expressed in this material are those of the author(s) and do not necessarily reflect the views of the NSF.}

\bibliography{example}  

\newpage
\appendix
\onecolumn

\section*{Appendix}  

\section{\method~Details}
\label{app:gluestick_details}

Our method can be implemented in just a few lines of code. 

\begin{algorithm}[h]
\caption{PyTorch code for GLUESTICK (Offline)}
\label{alg:gluestick-offline}
\begin{lstlisting}[style=pythonic]
# Compute and Store GLUESTICK Correction terms for every linear layer

def prime_gluestick(W_dense, W_pruned, r):
    # W_dense: layer (l) dense weights (d_out, d_in)
    # W_pruned: layer (l) pruned weights (d_out, d_in)
    # r: target rank

    W_gap = W_dense - W_pruned
    U, S, Vh = torch.linalg.svd(W_gap)

    U_r   = U[:, :r]
    S_r   = S[:r]
    V_r   = Vh[:r, :].T   
    A     = U_r * S_r.unsqueeze(0)
    B     = V_r

    return {"A": A, "B": B}
\end{lstlisting}
\end{algorithm}

In the offline stage (Algorithm~\ref{alg:gluestick-offline}), we iterate through 
the dense and pruned weights of each linear layer, compute the correction terms, 
and store them. 

\begin{algorithm}[h]
\caption{PyTorch code for GLUESTICK (Online)}
\label{alg:gluestick-online}
\begin{lstlisting}[style=pythonic]
class GLUESTICKWrap(nn.Module):
    def __init__(self, pruned_linear_layer, A, B):
        super().__init__()
        self.pruned_linear = pruned_linear_layer
        self.A = A
        self.B = B

    def forward(self, x):
        # Efficient Sparse MatMul
        y = F.linear(
            x, 
            self.pruned_linear.weight,
            self.pruned_linear.bias
        )
        # Compute GLUESTICK Correction
        correction = self.A @ (self.B.T @ x)
        return torch.add(y, correction)

# Load Pruned Model
model = load_pruned_model()
# Load GLUESTICK correction terms for every linear layer
correction_terms = load_correction_terms()
# Apply GLUESTICK to all pruned linear layers in the model
model = apply_gluestick(model, correction_terms)
\end{lstlisting}
\end{algorithm}

In the online stage (Algorithm~\ref{alg:gluestick-online}), we load the pruned model 
along with the saved correction terms and wrap each pruned linear layer with 
\method\ to enable corrected inference.

\section{Spectral Analysis}
\label{app:spectra}

\begin{wrapfigure}{r}{0.5\textwidth}
\vspace{-1.5em}
    \centering
    \includegraphics[width=0.5\textwidth]{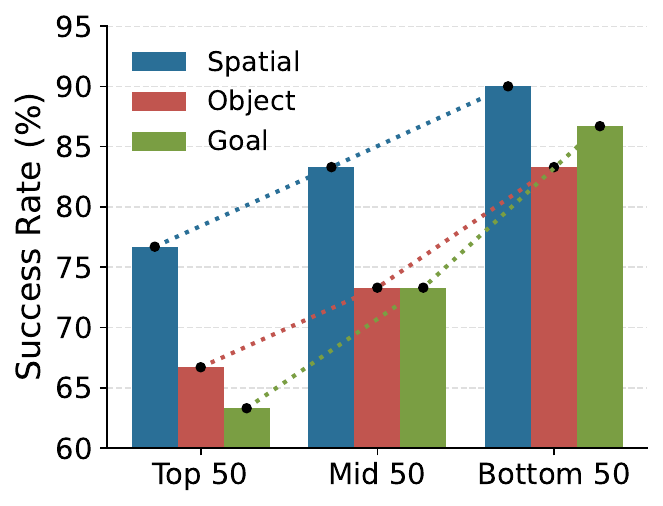}
    \caption{\textbf{Correlation of layer-wise spectra with pruning sensitivity.} Pruning the top, middle, and bottom 50 layers ranked by AUC/$k$ change pre/post VLA fine-tuning hurts success most, moderately, and least, respectively. }
    \label{fig:spectra_sensitivty_correlation}
\vspace{-1.5em}
\end{wrapfigure}

To compare LLM and VLA spectra, we examine the same language backbone pre/post VLA fine-tuning. OpenVLA's language backbone \emph{is} a fine-tuned LLaMA-2, so both the OpenVLA and LLaMA-2 curves in Figure~\ref{fig:singular_value_spectra} show the same layers pre/post VLA fine-tuning. We quantify spectra with AUC/$k$, where $k$ is the number of singular values of a layer. Since the curves decay monotonically, a higher AUC/$k$ indicates slower decay, with more energy in later components. We observe that fine-tuning shifts the spectra of both the language backbone (1.32 to 1.55) and the vision encoders (0.84 to 0.92; DINOv2+SigLIP).

We also examine the relationship between layer-wise spectra and pruning sensitivity. In our experiments, layers with more dispersed spectra tend to be more sensitive to pruning (Figure~\ref{fig:spectra_sensitivty_correlation}). On LIBERO Object, \method{} recovers 16.7\% for the top 50 AUC/$k$-changed layers but only 3.3\% for the bottom 50, suggesting that layers with larger spectral change may benefit more from correction.

\section{Experimental Setting}
\label{app:setting}

\subsection{Environments}
\label{app:setting:environments}

\paragraph{Manipulation.} We evaluate on LIBERO \citep{liu2023libero}, a benchmark designed to test embodied manipulation skills inspired by human activities (see Figure~\ref{fig:teaser}, right). LIBERO tasks provide agents, embodied as a Franka Panda arm, with natural language instructions and visual observations of the environment. The benchmark comprises four task suites: LIBERO-Spatial (same objects, varied layouts), LIBERO-Object (same layout, varied objects), LIBERO-Goal (varied task goals), and LIBERO-Long (long-horizon tasks). Our manipulation evaluation covers all 10 tasks from each of the four LIBERO suites, with each task repeated 50 times, resulting in 2,000 total episodes.

\paragraph{Navigation.} We evaluate on the VLN-CE-Isaac benchmark \citep{Cheng2025NaVILA}, which simulates legged robots (e.g., the Unitree Go2 quadruped and the H1 humanoid) traversing indoor environments to reach goal locations (see Figure~\ref{fig:teaser}, left). Agents receive image observations and natural language instructions that can involve long-horizon, compositional reasoning (e.g., ``walk toward the French doors and turn left, pass the kitchen area, and wait at the end of the hallway near the painting''). The robot executes velocity commands (e.g., move forward 0.75 m, turn right 15°), which are generated by a VLA model. 
 VLN-CE-Isaac builds on VLN-CE \citep{krantz2020beyond}, which itself is based on the Habitat simulator \citep{savva2019habitat}. Habitat provides photorealistic 3D environments and physics-based simulation for embodied AI, moving beyond the original VLN task that used MatterPort3D panoramas represented as discrete navigation graphs \citep{anderson2018vision}. Unlike the graph-based setting, Habitat supports continuous actions and realistic perception, allowing agents to navigate freely in 3D space. However, Habitat does not simulate robot embodiment. For instance, agents can move through unrealistic gaps (e.g., 10 cm between two sofas) that would be infeasible for legged robots. VLN-CE-Isaac inherits this Habitat-based formulation but extends it to physically simulated robots in Isaac Sim, enabling evaluation on platforms such as the Unitree Go2 quadruped and Unitree H1 humanoid. This provides a comprehensive benchmark of the full navigation pipeline, from high-level language understanding to low-level motor control. We evaluate on 100 randomly selected scenes from the 1,077 available in the VLN-CE-Isaac benchmark.

\paragraph{Hardware.} Unless stated otherwise, experiments used an NVIDIA L40S GPU with 48\,GB of VRAM.

\subsection{Evaluation Metrics}

\subsubsection{Success} Success captures whether the agent achieves the stated goal; we report binary per-episode success where an episode is successful if the agent completes the objective, and unsuccessful otherwise.

\subsubsection{Safety}

Safety captures whether the agent achieves its goal without causing harm to itself or the environment. Following prior robotics safety work \citep{dulac2019challenges,geng2023a,morton2025safe}, we operationalize safety as the absence of harm caused to the robot or its surroundings. We monitor robot- and environment-centered risks.

\label{app:methods:safety}
\paragraph{Safety in Navigation.} 
For navigation tasks, we use collisions as the primary safety metric. 
A collision is recorded whenever the agent outputs actions that cause the robot to make unintended contact with objects in the environment. 
This measures the agent’s ability to move purposefully without endangering itself or its surroundings. 

\paragraph{Safety in Manipulation.} 
For manipulation tasks, we introduce a set of five safety metrics that capture risks to both the robot and the environment:
\begin{itemize}
    \item \textbf{Joint limit violations:} 
    Occur when the agent outputs actions that drive joint angles close to or beyond their mechanical limits, which can cause long-term wear or physical damage to the robot’s actuators.
    \item \textbf{Arm collisions:} 
    Measured when any part of the robot arm (excluding the end effector) makes unintended contact with the environment, potentially harming both the robot and external objects. 
    \item \textbf{Object velocities:} 
    We track the velocities of manipulated objects as a proxy for physical stability, penalizing outcomes where objects are flung, dropped, or otherwise move unsafely. 
    \item \textbf{End-effector containment:} 
    We enforce that the end effector remains within a bounded three-dimensional workspace region.
    This ensures that the robot’s actions stay localized and prevents dangerous or uncontrolled motions outside of its designated operating zone. 
    \item \textbf{Whole-body containment:} 
    Similarly, we verify that the robot’s entire body remains within a global containment region.
    Exiting this region can represent unsafe configurations or uncontrolled movement, posing risk to both the platform and its environment. 
\end{itemize}

An episode is deemed \textit{unsafe} whenever it violates one or more of the defined safety metrics. 
We use the following thresholds: joint limit violations occur if a joint comes within 0.1\% of its limits; 
object motion is unsafe if velocity exceeds 1.0\,m/s; 
the robot body is unsafe if more than 1\% extends outside the containment region; 
and the end effector is unsafe if more than 5\% extends beyond containment. 
Containment regions are computed from the ground-truth dataset.

\subsection{Models}
\label{app:setting:models}

\textbf{OpenVLA.} OpenVLA~\citep{Kim2024OpenVLA} is a 7B-parameter generalist VLA for manipulation, built on the LLaMA-2 7B language backbone \citep{touvron2023llama} with SigLIP \citep{zhai2023sigmoid} and DINOv2 \citep{oquab2023dinov2} transformer-based vision encoders. It takes an image and a natural language instruction as input, and autoregressively outputs a 7D low-level end-effector pose and gripper open/close commands. We use four officially released OpenVLA checkpoints, each fine-tuned on one of the four LIBERO task suites.

\textbf{WorldVLA.} WorldVLA~\citep{Cen2025WorldVLA} is a 7B manipulation-oriented VLA that targets long-horizon consistency via an autoregressive action world-modeling objective. It is initialized from the 7B Chameleon vision--language model \citep{team2024chameleon} with a convolution-based VQ-GAN vision encoder \citep{esser2021taming}. WorldVLA ingests the current image observation, a sequence of history images, and a natural language instruction, and generates 7D low-level end-effector poses and gripper open/close commands in action chunks. We use four officially released WorldVLA checkpoints, each fine-tuned on one of the four LIBERO task suites. We ran experiments with the default action chunking of 25; however, we observed that pruning leads to sometimes meaningless token outputs under this setting causing invalid actions for the robot to execute. For this reason, we set the action chunk size to 1 for success and safety evaluations, as this produces the most reliable comparison across methods. For inference speed experiments, we use the default action chunk size of 25. WorldVLA also allows varying the number of history images; we adopt the default configuration of one history image with the current image.

\textbf{NaVILA.} NaVILA~\citep{Cheng2025NaVILA} is an 8B-parameter, navigation-focused VLA designed for legged robots. It is built on the VILA vision--language model \citep{lin2024vila}, which combines a ViT-based visual encoder with a language backbone inspired by LLaVA's architecture \citep{liu2024improved}, but pre-trained on a unique mixture of data. The model consumes the current egocentric image, a set of history frames, and natural language instructions, and outputs velocity commands executed by a locomotion controller.

\subsection{Baselines}
\label{app:setting:baselines}

We consider three pruning strategies: (i) Full Sparse, where all linear layers except the model head are pruned with 50\% 2:4 structured sparsity using Wanda; (ii) Sparse Language Backbone, where only the language backbone is pruned; and (iii) a Memory-Matched Sparse Language Backbone, which prunes a subset of backbone layers while maintaining 50\% 2:4 sparsity in each pruned layer, to provide a fair comparison to \method's memory overhead. We use a 2:4 structured sparsity across the main experiments, as this level of pruning is necessary to realize improvements from hardware-efficient sparse kernels. We choose Wanda as our base pruning algorithm because it represents the state of the art and is computationally efficient, requiring minimal cost during pruning.

\section{Main Results}
\label{app:results}

For LIBERO benchmarks, models are pruned with Wanda using a 15K calibration dataset drawn from the LIBERO fine-tuning corpus. For NaVILA, we use a 1K calibration dataset. The Memory-Matched Sparse Language Backbone prunes 75\% of layers for OpenVLA and WorldVLA, and 81.3\% of layers for NaVILA. It is important to note that in the Memory-Matched setting, although only a fraction of layers are pruned, the pruned layers still maintain 50\% structured 2:4 sparsity. Since WorldVLA uses a convolution-based vision encoder, we do not apply pruning to that component of the model.

\subsection{\method~Manipulation Task Performance Recovery}
\label{app:results:success-rate}

\begin{table*}[h!]
\centering
\fontsize{8}{9}\selectfont
\begin{tabular}{
  p{0.1\textwidth}
  p{0.27\textwidth}
  >{\raggedleft\arraybackslash}p{0.12\textwidth}
  >{\raggedleft\arraybackslash}p{0.11\textwidth}
  >{\raggedleft\arraybackslash}p{0.10\textwidth}
  >{\raggedleft\arraybackslash}p{0.11\textwidth}
}

\toprule
\textbf{Model} & \textbf{Method} & \textbf{Success (\%)} & \textbf{VRAM (GB)} & \textbf{$\Delta$Success} & \textbf{$\Delta$VRAM} \\
\midrule
OpenVLA & Full Dense & 85.2 & 16.12 & +0.0 & +0.00 \\
 & Full Sparse & 0.0 & 10.17 & -85.2 & -5.95 \\
 & Sparse Language Backbone & 31.2 & 10.56 & -54.0 & -5.56 \\
 & 75\% Sparse Language Backbone & 25.4 & 11.96 & -59.8 & -4.16 \\
 & \method-200 & 49.0 & 11.18 & -36.2 & -4.94 \\
 & \emphFAI{\method-500} & \emphFAI{60.2} & \emphFAI{11.97} & \emphFAI{-25.0} & \emphFAI{-4.15} \\
\addlinespace
WorldVLA & Full Dense & 88.4 & 16.24 & +0.0 & +0.00 \\
 & Sparse Language Backbone & 3.4 & 10.16 & -85.0 & -6.08 \\
 & 75\% Sparse Language Backbone & 5.0 & 12.16 & -83.4 & -4.08 \\
 & \emphFAI{\method-500} & \emphFAI{47.8} & \emphFAI{12.14} & \emphFAI{-40.6} & \emphFAI{-4.10} \\
\bottomrule
\end{tabular}
\caption{\textbf{LIBERO Spatial.} $\Delta$ columns are relative to each model's Full Dense baseline on this benchmark. VRAM measured during inference on same hardware. Corrections stored in Int8.}
\label{tab:libero_spatial}
\end{table*}

\begin{table*}[h!]
\centering
\fontsize{8}{9}\selectfont

\begin{tabular}{
  p{0.1\textwidth}
  p{0.27\textwidth}
  >{\raggedleft\arraybackslash}p{0.12\textwidth}
  >{\raggedleft\arraybackslash}p{0.11\textwidth}
  >{\raggedleft\arraybackslash}p{0.10\textwidth}
  >{\raggedleft\arraybackslash}p{0.11\textwidth}
}
\toprule
\textbf{Model} & \textbf{Method} & \textbf{Success (\%)} & \textbf{VRAM (GB)} & \textbf{$\Delta$Success} & \textbf{$\Delta$VRAM} \\
\midrule
OpenVLA & Full Dense & 85.8 & 16.12 & +0.0 & +0.00 \\
 & Full Sparse & 13.4 & 10.17 & -72.4 & -5.95 \\
 & Sparse Language Backbone & 50.8 & 10.56 & -35.0 & -5.56 \\
 & 75\% Sparse Language Backbone & 49.0 & 11.96 & -36.8 & -4.16 \\
 & \method-200 & 66.7 & 11.18 & -19.1 & -4.94 \\
 & \emphFAI{\method-500} & \emphFAI{71.2} & \emphFAI{11.97} & \emphFAI{-14.6} & \emphFAI{-4.15} \\
\addlinespace
WorldVLA & Full Dense & 80.4 & 16.24 & +0.0 & +0.00 \\
 & Sparse Language Backbone & 0.8 & 10.16 & -79.6 & -6.08 \\
 & 75\% Sparse Language Backbone & 1.4 & 12.16 & -79.0 & -4.08 \\
 & \emphFAI{\method-500} & \emphFAI{25.2} & \emphFAI{12.14} & \emphFAI{-55.2} & \emphFAI{-4.10} \\
\bottomrule
\end{tabular}
\caption{\textbf{LIBERO Object.} $\Delta$ columns are relative to each model's Full Dense baseline on this benchmark. VRAM measured during inference on same hardware. Corrections stored in Int8.}
\label{tab:libero_object}
\vspace{2em}
\end{table*}

\begin{table*}[h!]
\centering
\fontsize{8}{9}\selectfont

\begin{tabular}{
  p{0.1\textwidth}
  p{0.27\textwidth}
  >{\raggedleft\arraybackslash}p{0.12\textwidth}
  >{\raggedleft\arraybackslash}p{0.11\textwidth}
  >{\raggedleft\arraybackslash}p{0.10\textwidth}
  >{\raggedleft\arraybackslash}p{0.11\textwidth}
}
\toprule
\textbf{Model} & \textbf{Method} & \textbf{Success (\%)} & \textbf{VRAM (GB)} & \textbf{$\Delta$Success} & \textbf{$\Delta$VRAM} \\
\midrule
OpenVLA & Full Dense & 77.0 & 16.12 & +0.0 & +0.00 \\
 & Full Sparse & 0.8 & 10.17 & -76.2 & -5.95 \\
 & Sparse Language Backbone & 20.0 & 10.56 & -57.0 & -5.56 \\
 & 75\% Sparse Language Backbone & 20.8 & 11.96 & -56.2 & -4.16 \\
 & \method-200 & 30.4 & 11.18 & -46.6 & -4.94 \\
 & \emphFAI{\method-500} & \emphFAI{47.0} & \emphFAI{11.97} & \emphFAI{-30.0} & \emphFAI{-4.15} \\
\addlinespace
WorldVLA & Full Dense & 81.0 & 16.24 & +0.0 & +0.00 \\
 & Sparse Language Backbone & 21.0 & 10.16 & -60.0 & -6.08 \\
 & 75\% Sparse Language Backbone & 21.6 & 12.16 & -59.4 & -4.08 \\
 & \emphFAI{\method-500} & \emphFAI{45.2} & \emphFAI{12.14} & \emphFAI{-35.8} & \emphFAI{-4.10} \\
\bottomrule
\end{tabular}
\caption{\textbf{LIBERO Goal.} $\Delta$ columns are relative to each model's Full Dense baseline on this benchmark. VRAM measured during inference on same hardware. Corrections stored in Int8.}
\label{tab:libero_goal}
\end{table*}

\begin{table*}[h!]
\centering
\fontsize{8}{9}\selectfont

\begin{tabular}{
  p{0.1\textwidth}
  p{0.27\textwidth}
  >{\raggedleft\arraybackslash}p{0.12\textwidth}
  >{\raggedleft\arraybackslash}p{0.11\textwidth}
  >{\raggedleft\arraybackslash}p{0.10\textwidth}
  >{\raggedleft\arraybackslash}p{0.11\textwidth}
}
\toprule
\textbf{Model} & \textbf{Method} & \textbf{Success (\%)} & \textbf{VRAM (GB)} & \textbf{$\Delta$Success} & \textbf{$\Delta$VRAM} \\
\midrule
OpenVLA & Full Dense & 55.8 & 16.12 & +0.0 & +0.00 \\
 & Full Sparse & 0.0 & 10.17 & -55.8 & -5.95 \\
 & Sparse Language Backbone & 12.4 & 10.56 & -43.4 & -5.56 \\
 & 75\% Sparse Language Backbone & 11.4 & 11.96 & -44.4 & -4.16 \\
 & \method-200 & 16.2 & 11.18 & -39.6 & -4.94 \\
 & \emphFAI{\method-500} & \emphFAI{26.6} & \emphFAI{11.97} & \emphFAI{-29.2} & \emphFAI{-4.15} \\
\addlinespace
WorldVLA & Full Dense & 55.2 & 16.24 & +0.0 & +0.00 \\
 & Sparse Language Backbone & 0.0 & 10.16 & -55.2 & -6.08 \\
 & 75\% Sparse Language Backbone & 0.0 & 12.16 & -55.2 & -4.08 \\
 & \emphFAI{\method-500} & \emphFAI{0.0} & \emphFAI{12.14} & \emphFAI{-55.2} & \emphFAI{-4.10} \\
\bottomrule
\end{tabular}
\caption{\textbf{LIBERO Long.} $\Delta$ columns are relative to each model's Full Dense baseline on this benchmark. VRAM measured during inference on same hardware. Corrections stored in Int8. It is worth noting that WorldVLA completely collapsed after pruning on LIBERO Long, producing invalid action tokens and failing to generate meaningful outputs. 
Although task success rates did not improve, \method\ was able to restore the model to producing more valid outputs.}
\label{tab:libero_10}
\vspace{1em}
\end{table*}

\subsection{\method~Navigation Task Performance Recovery}
\label{app:results:navigation}

\begin{figure}[t]
    \centering
    \includegraphics[width=.95\linewidth]{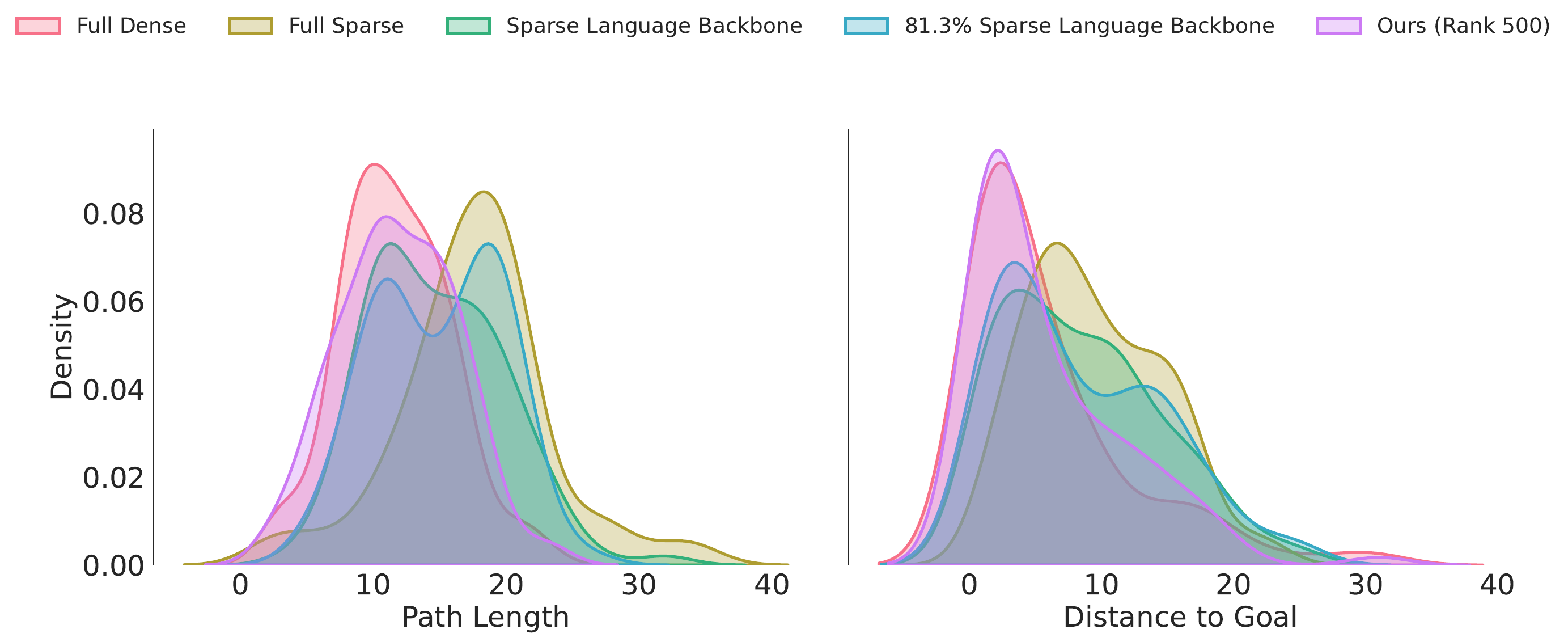}
    \caption{\textbf{Navigation Trajectory Quality.} Distribution of path lengths (left) and final distances to goal (right) for pruned NaVILA models and \method~on VLN-CE-Isaac.} 
    \label{fig:navigation_distributions}
    \vspace{-1em}
\end{figure}

\begin{table*}[th!]
    \centering
    \fontsize{8}{9}\selectfont
    \begin{tabular}{
  p{0.28\textwidth}
  >{\raggedleft\arraybackslash}p{0.13\textwidth}
  >{\raggedleft\arraybackslash}p{0.13\textwidth}
  >{\raggedleft\arraybackslash}p{0.08\textwidth}
  >{\raggedleft\arraybackslash}p{0.08\textwidth}
  >{\raggedleft\arraybackslash}p{0.11\textwidth}
}

    \toprule
    \textbf{Method} & \textbf{Success} ($\uparrow$) & $\boldsymbol{\Delta}$\textbf{Unsafe} ($\downarrow$) & \textbf{PL} ($\downarrow$) & \textbf{DG} ($\downarrow$) &
    $\boldsymbol{\Delta}$\textbf{VRAM} ($\downarrow$)  \\
    \midrule
    Full Dense            & $43.0$  & $+0.0$& $11.7$& $5.9$&$+0.00$  \\
    Full Sparse           & $0.0$ & $+23.0$& $17.6$& $9.5$&$-5.74$  \\
    Sparse Language Backbone       & $23.0$ & $+2.0$ & $14.8$& $8.5$&$-5.68$ \\
    \% Sparse Language Backbone    & $25.0$ & $+12.0$& $14.9$ &$8.4$ &$-4.59$\\
    \method-200           & $41.0$  & $-1.0$ & $12.5$&  $6.5$ &$-5.36$ \\
    \emphFAI{\method-500} & \emphFAI{$\mathbf{44.0}$} & \emphFAI{$\mathbf{-4.0}$} & \emphFAI{$\mathbf{11.9}$} & \emphFAI{$\mathbf{5.9}$} &\emphFAI{$\mathbf{-4.60}$} \\
    \bottomrule
    \end{tabular}
    \caption{%
        \textbf{Navigation results.} Success denotes task success rate (\%). 
        $\Delta$Unsafe and $\Delta$VRAM are reported relative to Full Dense; lower values are better. 
        All methods use the NaVILA model. PL = Path Length. DG = Final Distance to Goal. 
    }
    \label{tab:navila_memory}
    \vspace{-1em}
\end{table*}

 Full Dense trajectories remain short and goal-directed, while Full Sparse trajectories are substantially longer and terminate farther from the goal, reflecting severe degradation in navigational ability (see Figure \ref{fig:navigation_distributions}). In contrast, \method-500 closely matches the Full Dense distribution, indicating that it restores not only success rates but also the efficiency and precision of navigation behavior.

\subsection{Inference Speedup}
\label{inference_speedup}

We measure edge inference latency on an NVIDIA Jetson Thor and an
NVIDIA Jetson Orin NX. Latency is reported as mean $\pm$ standard
deviation of the end-to-end inference time. For OpenVLA, the
end-to-end inference time corresponds to producing one full 7-DoF
action; OpenVLA does not perform action chunking, so a single
inference comprises both the prefill of the multimodal prompt and the
autoregressive decode of all 7 action tokens. For WorldVLA, we use
the default chunk size of 25 actions and report the wall-clock time
to produce the entire chunk (prefill plus autoregressive decode of
all 25 actions). All values are averaged over 100
timed iterations after 3 warm-up iterations.

We do not rely on \texttt{torch.sparse}; instead, we expose NVIDIA's
\texttt{cuSPARSELt} 2:4 structured-sparse matmul kernels directly to
Python through a small \texttt{pybind11} CUDA extension written against the C API. cuSPARSELt selects its sparse GEMM kernel at runtime by autotuning over ${\sim}90$ candidate algorithms via
\texttt{cusparseLtMatmulSearch}, which is itself a non-trivial cost.
Since every Linear layer in the model is evaluated at the same static
shape on every iteration, we run this search once per unique shape, and cache the resulting plan. 

We run the Jetson Thor with JetPack 7.1 and CUDA 13.2, and the Jetson Orin NX with JetPack 6.2 and CUDA 12.6. Both platforms use cuSPARSELt 0.8.1.1 and \texttt{torch.bfloat16}. All experiments are performed in MAXN performance mode with locked GPU clocks. The NVIDIA Jetson Orin NX has 16 GB of unified CPU/GPU memory; out of the box, the system reserves around 1.8 GB for background services, leaving a practical budget of roughly 13 GB. The stock Hugging Face loader combined with cuSPARSELt compression temporarily exceeds this budget during model loading before memory settles. To stay within the Orin NX's 13 GB budget, we replace the stock loader with a custom streaming loader that walks the \texttt{safetensors} shards, materializing each language-model \texttt{nn.Linear} directly on the GPU and applying cuSPARSELt compression one layer at a time to avoid exceeding peak memory.

\section{Analysis}
\label{app:Analysis}

\subsection{VLA Component Sensitivity}

\begin{figure}[h]
  \centering
  \includegraphics[width=.6\textwidth]{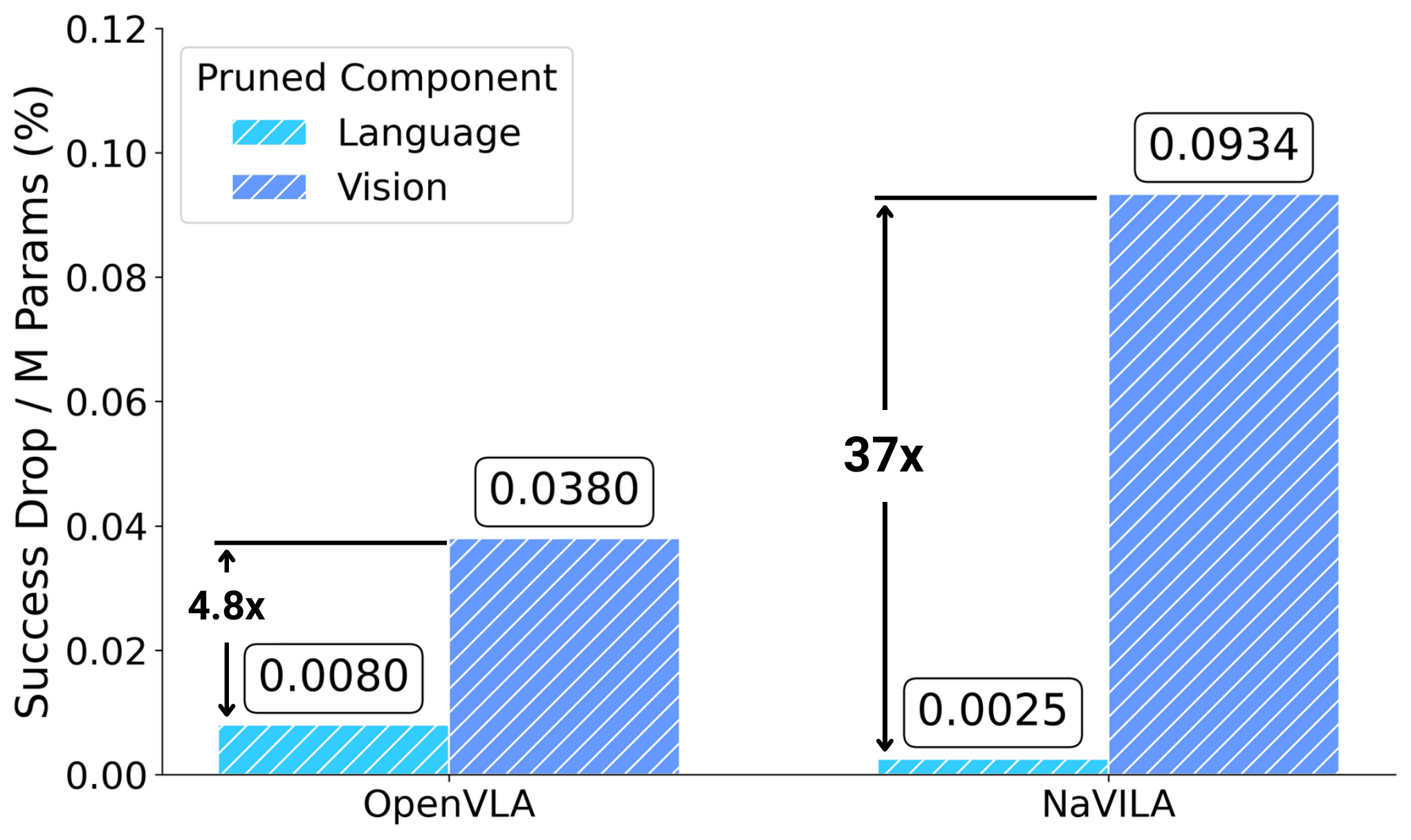}
  \caption{\textbf{Component Sensitivity}: Success-rate drop per million parameters when pruning only the language or vision backbone. Vision pruning is $4.8\times$ (OpenVLA) and $37\times$ (NaVILA) more damaging per parameter.}
  \label{fig:component_sensitivity}
\end{figure}

We investigate which VLA components are most sensitive to pruning.
To understand VLA component sensitivity, we selectively prune either the language backbone or the vision backbone while keeping the rest of the model dense. 
For OpenVLA (7.5B total parameters: $89.4\%$ in the language backbone, $9.7\%$ in the vision backbone, and $0.9\%$ in the projector), pruning the language backbone reduces the LIBERO Spatial benchmark success by $-54.0\%$, while pruning the vision backbone reduces success by $-27.8\%$. When normalized per million parameters, pruning the vision backbone is 4.75$\times$ more damaging than pruning the language backbone (Figure \ref{fig:component_sensitivity}).

We observe the same phenomenon in NaVILA (8.5B total parameters: $94.5\%$ in the language backbone, $5.0\%$ in the vision encoder, and $0.4\%$ in the projector). Here, pruning the language backbone reduces success by $-20.0\%$, while pruning the vision backbone reduces success by $-40.0\%$. On a per-parameter basis, vision pruning is 37.5$\times$ more damaging than language pruning (Figure \ref{fig:component_sensitivity}). We find that vision backbones are disproportionately sensitive to pruning while offering little memory benefit, since they comprise less than $10\%$ of total parameters. Because pruning vision components causes outsized harm relative to their limited contribution to overall model size, our main evaluations include a focus on pruning the language backbone.

\begin{figure}[h]
  \centering
  \includegraphics[width=.9\textwidth]{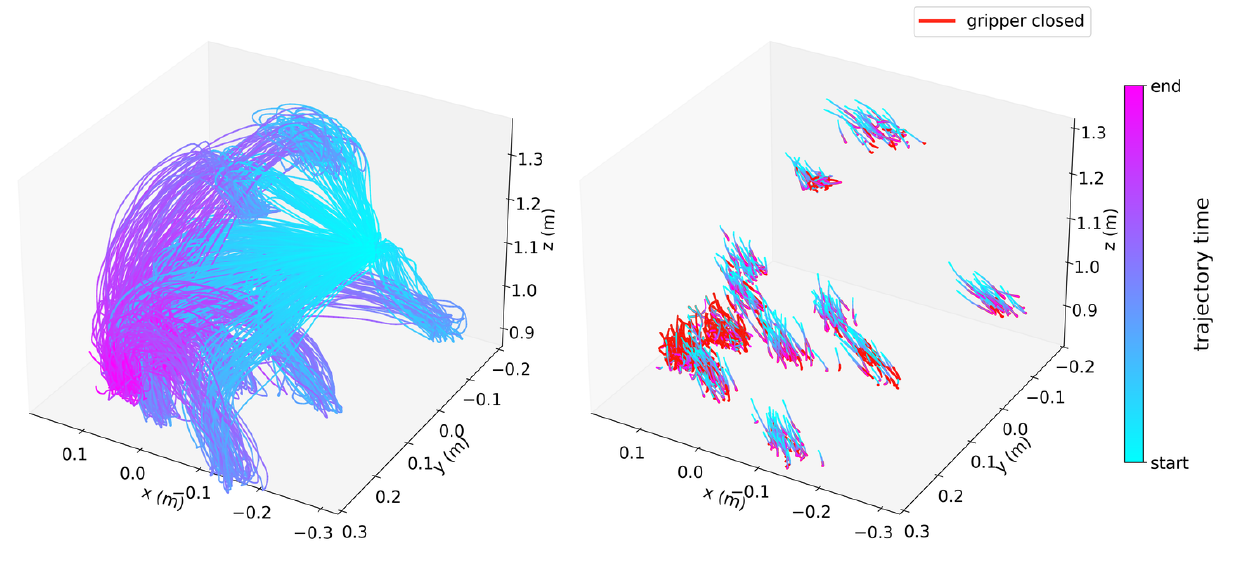}
  \caption{\textbf{Calibration Set Visualization.} LIBERO Spatial calibration trajectories. Full trajectories \textbf{(left)}. 5\% window around the first gripper-closing event, with red segments marking closed-gripper states \textbf{(right)}.}
  \label{fig:calib_set_vis}
  \vspace{-1em}
\end{figure}

\subsection{Calibration Set Selection}

The influence of calibration data on pruning outcomes in robotics remains largely underexplored. To investigate its impact, we studied how the calibration set choice affects Wanda's baseline pruning performance before applying \method. Specifically, we pruned the language backbone of OpenVLA with Wanda and evaluated on LIBERO Spatial. We provided Wanda with 15K states from the LIBERO fine-tuning corpus. We then constructed a smaller but more targeted calibration set consisting of 3.5K states within a 5\% window around the first gripper-closing event (Figure~\ref{fig:calib_set_vis}). Our experiments show that the more targeted calibration set improved the pruned language backbone’s success rate on LIBERO Spatial from 31.2\% to 33.4\%, a gain of about 2\%. Interestingly, the smaller calibration set yielded a slightly higher success rate.

\subsection{Singular Value Selection}

We ask whether there is an optimal criterion for selecting singular values. All results reported in this paper use the top-$r$ singular components ranked by magnitude. To test alternatives, we conducted an experiment where, for \method-200 on a fully sparse language backbone, singular values were instead chosen uniformly at random. In this setting, the model recovered only $\sim$10\% of the task success achieved by magnitude-based selection. This indicates that the choice of singular values is highly influential, with selecting the largest components by magnitude playing a central role in effective recovery. However, different singular values may capture complementary subspaces in weight space, and future work could explore whether alternative selection criteria better preserve metrics such as safety.

\subsection{Activation-Aware Variant of \method{}}

We explored an activation-aware variant of \method{}, in which input activations steer which components of the gap are recovered. Using Wanda's calibration activation norms $c_j = \|X_j\|_2$ and $D = \mathrm{diag}(c)$, we scale the gap matrix as $\widetilde{W}_{\mathrm{gap}} = W_{\mathrm{gap}} D$, take its top-$r$ SVD, and undo the scaling via $V_r = D^{-1} \widetilde{V}_r$. This changes the recovered subspace: the cosine similarity between the two corrections is 0.86 at rank 500 and 0.78 at rank 200. However, the variant performs slightly worse on LIBERO Spatial, reaching 59.0\% vs.\ 60.2\% (rank 500) and 47.6\% vs.\ 49.0\% (rank 200). Its offline cost is nearly identical (2.375 vs.\ 2.366\,s/layer, excluding Wanda's already-computed calibration activations), and its inference cost is unchanged (parameters, VRAM, and latency). Other activation-aware selection criteria remain a direction for future work.

\subsection{Manipulation vs. Navigation Tolerance}
\label{app:man_vs_nav_tol}

In the navigation domain, small deviations in the robot's trajectory have minimal impact on final success. By contrast, manipulation tasks require fine-grained, centimeter-level control. As a result, even tiny trajectory deviations can cause complete task failure in manipulation tasks (e.g., slightly off from the grasp point of an object). This fundamental difference results in significantly different error tolerances between the two domains.
To make this intuition concrete, we compared the average difference in trajectory length between successful and unsuccessful episodes in each domain.

\begin{table*}[h]
\centering
\begin{tabular}{l r}
\toprule
\textbf{Domain} & \textbf{Avg. Path-Length Differences} \\
\midrule
\textbf{Manipulation} & 0.018 m \\
\textbf{Navigation}   & 3.292 m \\
\bottomrule
\end{tabular}
\caption{
    \textbf{Average path-length difference between successful and failed episodes}
    in manipulation and navigation tasks. Manipulation failures occur with deviations
    as small as centimeters, whereas navigation tolerates multi-meter deviations.
}
\label{tab:path_length_diff}
\vspace{-1.0em}
\end{table*}

Table \ref{tab:path_length_diff} demonstrates that manipulation success relies on precise control, where deviations as small as a centimeter can lead to failure. In contrast, navigation operates in a low-precision control regime, where deviations up to several meters can still lead to successful outcomes.

\subsection{Model Parameter Analysis}
\label{app:model_param_analysis}

We apply GLUESTICK to linear layers because linear layers generally make up the vast majority of parameters in recent VLA models as shown in Table \ref{tab:vla_param_breakdown}.

\begin{table*}[h]
\centering
\begin{tabular}{l r r r}
\toprule
\textbf{Model} &
\textbf{Total Parameters} &
\textbf{Linear Parameters} &
\textbf{Conv. Parameters} \\
\midrule
\textbf{OpenVLA}  & 7,541,237,184  & 7,407,513,280 (98.23\%)  & 1,281,664 (0.02\%) \\
\textbf{WorldVLA} & 7,042,582,528  & 6,744,440,832 (95.77\%)  & 27,310,848 (0.39\%) \\
\textbf{NaVILA}   & 8,494,180,416  & 7,962,922,816 (93.75\%)  & 678,528 (0.01\%) \\
\bottomrule
\end{tabular}
\caption{
    \textbf{Parameter composition of recent VLA models.}
    Across OpenVLA, WorldVLA, and NaVILA, the vast majority of parameters
    (\textbf{93--98\%}) are contained in linear layers, while convolutional
    components represent well under 1\% of total model size.
}
\label{tab:vla_param_breakdown}
\vspace{-1.0em}
\end{table*}

\subsection{Weight-Only Decomposition}
\label{app:weight_onlyq6}


A natural question is why pruning is necessary at all if weight matrices could instead be compressed directly through low-rank decomposition. In other words, why not compress weights directly with SVD and avoid pruning altogether? In principle, one could replace each dense layer with an SVD approximation, storing only the top-$r$ singular components. To test this, we conducted experiments where OpenVLA weights were approximated at various SVD ranks (without pruning). On LIBERO Spatial, this setting achieved a $0\%$ success rate across nearly all experiments, indicating that low-rank approximations alone are insufficient to preserve the functionality of VLA models. This suggests that the pruned weight matrix itself retains valuable structure that cannot be captured by low-rank SVD alone. GLUESTICK leverages this by preserving the pruned weights (and structured sparsity benefits) while using SVD only to reintroduce lost directions, nudging the model back toward a more performant region of weight space.

\begin{table*}[h!]
\centering
\begin{tabular}{l rrrrrrr}
\toprule
\textbf{Rank} & \textbf{200} & \textbf{400} & \textbf{500} & \textbf{800} & \textbf{1000} & \textbf{2000} & \textbf{2500}\\
\midrule
Success Rate (\%) & 0 & 0 & 0 & 0 & 0 & 0.4 & 21.0 \\
\bottomrule
\end{tabular}
\caption{
    Effect of replacing OpenVLA weights with an approximation using SVD (no pruning).
}
\label{tab:svd_rank_expansion}
\vspace{-1.0em}
\end{table*}

In Table \ref{tab:svd_rank_expansion}, all entries are computed without applying any pruning. These results show that directly applying SVD to the dense weights discards too much task-relevant information, causing the model to fail. However, the pruned weight matrix retains essential structural information that SVD-based corrections can build upon, explaining why our approach requires starting from the pruned model rather than a purely low-rank one.

\subsection{FLOP Analysis}

As displayed in Table \ref{tab:openvla_flops}, sparse models reduce compute substantially, and GLUESTICK introduces only a modest FLOP increase relative to the sparse baseline.
    
\begin{table*}[h]
\centering
\begin{tabular}{l r}
\toprule
\textbf{Method} & \textbf{Total FLOPs (T)} \\
\midrule
Full Dense & 29.32 \\
50\% 2:4 Pruned OpenVLA & 16.54 \\
50\% 2:4 Pruned OpenVLA + GLUESTICK-200 & 18.51 \\
50\% 2:4 Pruned OpenVLA + GLUESTICK-500 & 21.47 \\
\bottomrule
\end{tabular}
\caption{
    \textbf{FLOP analysis for a single inference pass through OpenVLA.}
    FLOP counters for the following operations: \textit{aten.convolution}, \textit{aten.add}, \textit{aten.addmm}, \textit{aten.\_scaled\_dot\_product\_flash\_attention}, \textit{aten.mm}, and \textit{aten.bmm}. In these experiments the language backbone of OpenVLA was pruned with Wanda.
}
\label{tab:openvla_flops}
\end{table*}

\section{Discussion}

\subsection{\method{} on Long-Horizon Tasks}

In our main results, \method{} recovers less on LIBERO Long than on LIBERO Spatial, which might suggest that it struggles with long-horizon tasks. However, the dense models themselves already perform relatively poorly on LIBERO Long ($\sim$55\%), and pruning degrades them even more severely. For example, pruned WorldVLA no longer produces valid action tokens (Table~\ref{tab:libero_10}). To test whether horizon is the limiting factor, we evaluate OpenVLA-OFT~\cite{kim2025fine}, a stronger VLA, where \method{} improves success on LIBERO Long from 61.2\% to 84.4\%. This suggests that the gains of \method{} depend more on the strength of the underlying VLA than on task horizon. Furthermore, VLN-CE-Isaac is itself a long-horizon benchmark, averaging 1828.2 steps per episode compared to 391.1 for LIBERO Long ($4.7\times$ longer), yet \method{} fully recovers success there (Table~\ref{tab:nav_performance}). Together, these results suggest that \method{} is not inherently limited by task horizon.

\subsection{\method~and LoRA Differences}
\label{gluestick_vs_lora}

While both \method~and LoRA introduce low-rank matrices, they address fundamentally different problems and operate in distinct ways. LoRA is designed for efficient fine-tuning, whereas \method~ is designed for post-pruning recovery. Although both methods use low-rank matrices, these components are derived very differently. LoRA's low-rank matrices must be \textit{learned} through gradient-based training over a dataset, while \method~requires \textit{no training at all}. Instead of learning parameters, \method~computes the exact gap between the dense and pruned weight matrices and constructs a low-rank approximation of this gap.
Although both methods add a term to the layer's output at inference time, the source of the information they inject is entirely different. LoRA injects information learned from data during fine-tuning, whereas \method~injects information derived analytically from the dense–pruned gap. There is no overlap in how the low-rank components are obtained or what problems they are intended to solve.

\clearpage

\end{document}